\documentclass[preprint,12pt]{elsarticle}

\biboptions{sort&compress}

\usepackage[utf8]{inputenc}
\usepackage{amsmath,amssymb,amsfonts}
\usepackage{booktabs}
\usepackage{array}
\usepackage{multirow}
\usepackage{hyperref}
\usepackage{url}
\usepackage{longtable}
\usepackage{pdflscape}
\usepackage{graphicx}
\usepackage{subcaption}
\usepackage{gensymb}

\begin{document}

\begin{frontmatter}

\title{FEA solver integration of a composite World Model with a Spiking Neural Network memory Block for computationally efficient modelling of defects and variabilities}

\author[1]{Bassam El Said\corref{cor1}}
\ead{bassam.elsaid@bristol.ac.uk}

\author[2,3]{Adam Sobey}

\cortext[cor1]{Corresponding author}

\affiliation[1]{organization={Bristol Composites Institute,
                              University of Bristol},
                city={Bristol},
                country={United Kingdom}}

\affiliation[2]{organization={University of Southampton},
                city={Southampton},
                country={United Kingdom}}

\affiliation[3]{organization={Alan Turing Institute},
                city={London},
                country={United Kingdom}}

\begin{abstract}
Each fiber-reinforced composite structure has a unique material architecture. The manufacturing methods used and the presence of defects lead to local variations which require material-scale models that can resolve the response in real time at every location in the structure. Concurrent multi-scale schemes such as Finite Element Squared (FE²) provide this capability, but at a prohibitive computational cost for large structures as the number of integration points and time increments scales with size. Replacing the material-scale models with surrogates has been studied extensively to mitigate these computational limitations. A surrogate model must represent an arbitrary local architecture including defects, track the irreversible material state over a number of increments that is not known in advance, remain stable when driven recursively by the surrogate’s own predictions, and operate within the memory available at every integration point of a structural model. In this paper, a Composites World Model based on Multi-Modal Spiking Transformer (MMST) architecture is designed to meet these requirements. The architecture combines a de-noising $\beta$-Variational Auto-encoder that learns the material architecture including defects, a Spiking Neural Network (SNN) memory block that efficiently tracks the long-term internal state of the material, and transformer blocks that learn the short-term applied strain state. The system is trained in three stages: material architecture encoding, base training on short time windows, and recursive fine-tuning with scheduled sampling to suppress cumulative error drift. The trained model is disassembled at inference and coupled to a non-linear FE solver, enabling structural scale simulation of composites containing embedded defects. Benchmarks against experimental results for IM7\slash8552 laminates under tension, compression and transverse shear, for pristine and wrinkled composites, show that the SNN memory block successfully tracks the material internal state and results in prediction errors between 0.4\% and 4.8\%.

\end{abstract}


\end{frontmatter}

\section{Computationally efficient damage prediction of composite materials under quasi-static loading}
\label{sec:intro}

Efficient and safe composite design is dependent on failure prediction. The mechanical response: stiffness, damage initiation and damage progression \citep{lomov2001textile,desplentere2005microct,selvaraj2023multiscale}, are entirely determined by the internal architecture of the material. However, accurate and reliable numerical simulation of composite material behaviour is a challenging task due to a number of factors relating to how these materials are manufactured. During manufacturing, the internal architectures deform to conform to the structural geometry, leading to locally varying architectures and material properties. Additionally, manufacturing processes often introduce localized defects (wrinkles, gaps and voids), adding a further layer of property variability which can be stochastic in nature \citep{elsaid2016multiscale,hamila2008simulations,mesogitis2014uncertainty,croft2011experimental,heinecke2019manufacturing}. The material properties measured independently from the structure and manufacturing processes do not describe how the material will behave as part of a structure, since their properties do not include the impact of the geometric features and presence of defects and variabilities \citep{talreja2024failure,wisnom1996size}.

A second level of complexity originates from the multi-scale nature of the materials' mechanical response. Damage events often initiate at the Fibre/Matrix scale, as a result of the interaction between the architectural features and the applied loading conditions \citep{obert2014micro,elsaid2018iterative}. An example is transverse cracks appearing in a matrix pocket inside a 3D woven composite. These material-scale damage events do not correlate directly with structural-scale performance \citep{harris2003advanced}, but as loading increases these small cracks progress and interact gradually leading to structural scale failure\citep{pham2019discrete,kashtalyan2013predicting}. The separation between material-scale and structure-scale damage events means that to quantify structural failure behaviour in composites it is essential to observe the failure progression at the material architecture length scale for a period of time. Predicting damage initiation based on failure of an individual constituent (fibre and matrix)  will greatly underestimate a composite structure load carrying capacity.
This challenge is a multi-scale problem, where the complete response of a structure is a function of the structural-scale loads and damage progression on the material-scale. This progression is non-linear in nature, due to the stress distribution at the material-scale as a result of damage progression, and the associated change in structure-scale loading condition.

Currently, capturing the history dependent non-linear behaviour of hierarchical materials requires time-dependent FEA simulations using
phenomenological models based on the material constituents properties and observed damage modes
\citep{puck1998failure,ladeveze2000mesomodel}. These models, while accurate, are computationally prohibitive for structural-scale
simulations. It necessitates high fidelity simulation at every location in a structure. For large scale structures such as aircraft
wings (60--80\,m) and wind turbine blades (60--100\,m), a high-fidelity simulation might require millions of evaluations of
the material constitutive law for each time iteration.  This makes the computational expense of these models prohibitive as a design tool \citep{bessa2017framework,fritzen2016finite}. Concurrent multi-scale schemes such as Finite Element Squared (FE²) \citep{feyel2000fe2} provide this capability, but at a cost that scales with the number of integration points and time increments, since a Representative Volume Element (RVE) model has to be solved for each integration point. This computational cost becomes prohibitive for large structures.

Replacing the material-scale models with surrogate models has been studied extensively to mitigate these computational limitations. However, these surrogates must satisfy requirements imposed by the FE solver beyond accurately representing the material response. Here, the surrogate model must represent an arbitrary local architecture including defects, track the irreversible material state over a number of increments that is not known in advance, remain stable when driven recursively by the surrogate’s own predictions, and operate within the memory available at every integration point of a structural model.\citep{bessa2017framework,kim2021deep,lim2021extrapolative}. This challenge has been addressed previously using Recurrent Neural Networks (RNN) \citep{ghane2023recurrent,kovacs2025physically,Maia2023}, which are designed to model time series data based on internal memory mechanisms. RNN are associated with high computational expense in both training and inference as each neuron has a memory. In practice, RNN-based surrogates have been demonstrated predominantly on academic-scale problems \citep{ghavamian2019accelerating}, and their behaviour when deployed at every integration point of a structural model has received little attention. Integrating RNN surrogates with FE solvers will lead to prohibitive memory requirements. This arises from the need to maintain the history of each integration point in the form of a complete copy of an RNN network's internal memory unique to each integration point. While the memory cost of an individual copy of an RNN can be small, this expands rapidly as the number of integration points reach 100 of thousands or millions, as is the case for industrial scale FE models.  

Another limitation of history dependent material surrogate models is that they have been shown to suffer from error drift over long predictions. This means that for long FEA simulations, errors will start accumulating as the simulation progress often leading to erroneous results for failure prediction\citep{elsaid2023predicting}. Here, small stress-prediction errors will lead to  errors in the subsequent FE steps, Internal Force Vector calculations and global equilibrium, which in turn will lead to further errors in strain and stress predictions for the next time step. 

Transformers are an alternative to RNNs, which has seen considerable success in Large Language Models (LLMs), and have success in materials application \citep{Methods2024, Rane2023}. However, Transformers are difficult to reconcile with the incremental nature of multi-scale solvers. Modern LLMs support context lengths of $10^5$ to $10^6$ tokens, but that capability is provisioned for a single instance evaluated once per query. In a multi-scale FEA simulation, the model is instead evaluated at every integration point at every increment, and a context must be maintained for each integration point independently. This is further complicated by the fact that the number of increments and the strain step are not known in advance, since FE solvers adapt the time step to maintain stability. Retaining this information, or the full stress-strain history for each integration point, will lead to the same excessive memory requirements observed for RNN. These issues have been addressed in long-context LLMs by reducing the unbounded history into a fixed-length context \citep{bulatov2022recurrent,munkhdalai2024leave}. World Models are an example of such architectures which employ an encoder/memory/predictor model \citep{ha2018world,hafner2019learning}. Under this modeling paradigm, the history is encoded and then submitted to a memory module which stores the “world state” and provides a context for the transformer prediction.

In this paper, a novel Multi-modal Transformer Architecture with a Spiking Neural Network memory block (MMST) is proposed for multi-scale
simulation of hierarchical materials. Inspired by World Model type architectures, the proposed model has the ability to learn the material architecture, history dependencies, and mechanical response. The use of a memory block in this architecture gives the overall system the ability to integrate seamlessly with a non-linear FEA solver without being burdened by the transformer context size. To achieve robust performance in an FEA simulation environment, the full system is trained in three subsequent stages: Material Architecture Pretraining, Base Training (short-term) and Recursive Training (long-term). Finally, the paper covers the efficient integration of the trained system in the time integration loop of FEA solvers, with comparisons against experimental results of composite material with defects and variabilities.

\section{Deep Learning Powered Multi-Scale Modelling}
\label{sec:dl_multiscale}

\subsection{Multi-scale Modelling Requirements for Deep Learning Models}
\label{sec:requirements}

Numerical approaches for nonlinear multi-scale modelling of strongly
heterogeneous materials often employ computational homogenisation,
with FE$^2$ \citep{feyel2000fe2}  based
approaches being the most common. This family of approaches link a
structural-scale model with a number of material-scale
models. Each material-scale model represents the material
architecture in the vicinity of the macro-scale integration point.
FE$^2$ approaches involves two main stages: Homogenisation and Localisation.

\paragraph{Homogenisation} Each macro-scale integration point is
assigned a material constitutive behaviour based on the local
underlying architecture. This constitutive behaviour could take the
form of a constitutive material tensor or stresses calculated at the
integration point. Representative Volume Element (RVE) models are
built to represent the local material architecture and it is impact on the constitutive behaviour. The RVE models
are solved under the applied strains to calculate full field
stress/strains. Most commonly used approaches involve 1st Order
Homogenisation (1OH), which links an RVE to a macro-scale integration
point. The full field stresses are homogenised following,
\begin{align}
\Sigma_X &= \frac{1}{V}\int_V \sigma(X,y)\,\mathrm{d}V,
\label{eq:homog_stress}\\
E_X &= \frac{1}{V}\int_V \varepsilon(X,y)\,\mathrm{d}V.
\label{eq:homog_strain}
\end{align}
Where $\Sigma_X$ are the macro-scale stresses defined in the
macro-model coordinates $X$ and $\sigma(X,y)$ are the material-scale
stresses defined in the material-model coordinates $y$. $E_X$ are the
macro-scale strains and $\varepsilon(X,y)$ are the material-scale strains
and $V$ is the RVE volume. Using this homogenisation, the RVE
response, which is defined by 3D stress and strain fields, is
compressed into average stress/strain at the macro-scale integration
point. Here, it is worth noting that more recently Second Order
Computational (2OH) homogenisation has grown in popularity for
modelling strongly heterogeneous materials
\citep{kouznetsova2002multi,hii2022kinematically}. In 2OH
approaches, the average strains/stresses and their spatial gradients
are passed between the length scales, as opposed to the averages
only for 1OH. While these approaches are very robust to
scale-separation challenges, they are associated with an
exceptionally high dimension strain space, which poses  a unique
challenge for surrogate modelling, and will be out of scope for this
work.

\paragraph{Localisation} forms the second step in the FE$^2$ loop.
At each time step, the macro-scale model is solved using the homogenised
properties mapped from the RVE models to integration points. The
macro-scale strains are extracted and then applied as boundary
conditions to the RVE models. These material-scale models are then
solved and homogenised to predict the constitutive behaviour for the
next macro-scale time-step.

\begin{figure}
\centering
\includegraphics[width=0.9\textwidth]{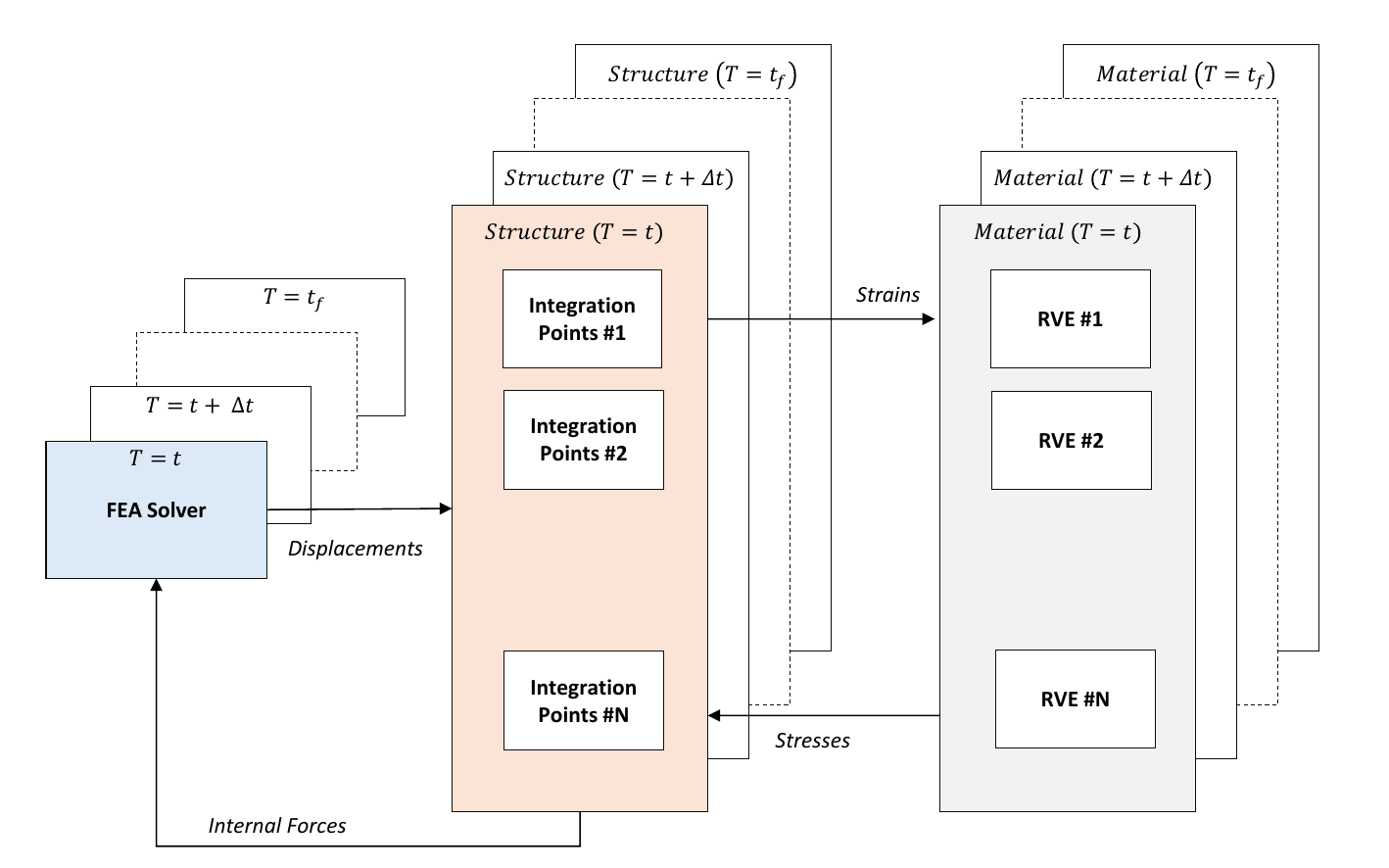}
\caption{A schematic of information flow in a concurrent multi-scale solver between material
and structural scales}
\label{fig:fe2flow}
\end{figure}

Running the full FE$^2$--1OH loop, involves solving the macro-scale
model and every associated RVE model at each FEA solver time step,
followed by computational homogenisation and information exchange
between the length scales. An overview of the FE$^2$ loop is shown in
Figure~\ref{fig:fe2flow}. Due to the number of models involved in
this approach, this process is computationally prohibitive for any
practical scale models. Consequently, FE$^2$ models use is primarily
limited to benchmarking multi-scale modelling techniques. All
practical applications of FE$^2$ involves some form of approximation
of the full loop. Of interest for this work, is the family of
approaches where the RVE models are replaced with a time-dependent
surrogate\citep{ElFallakiIdrissi2024,Ghane2024}. Here, the surrogate model replaces the RVE models in the
online FE$^2$ loop. To fulfil this role, such surrogates need two key capabilities :
\begin{itemize}
\item Represent the response of the underlying material architecture.
For example, in a composite, a surrogate should be able to predict
the material response including stiffness, damage initiation and
failure, while taking into account the local lay-up and the
presence of any defects and variabilities.
\item Track the material history, this includes the applied strain
history and the associated material state including damage initiation
and progression.
\end{itemize}

These functionalities are carried out per integration point, which is
a considerable computational challenge for most deep learning
models due to memory cost. The deep learning approach
proposed in this paper was constructed from the ground up for FEA
solver integration. Consequently, memory management requirements and time
integration challenges were given priority during model architecture
and training.

\subsection{Spatial / Temporal Deep Learning System Architecture}
\label{sec:architecture}

Multi-scale modelling of progressive damage in a hierarchical material is a combined temporal/spatial learning task, 
where the model needs to learn both the material architecture and material state. To tackle both learning tasks and develop a memory efficient deep learning approach, the proposed system is constructed from a number of modular blocks, each focusing on one task. The proposed architecture is inspired by World Models, and builds on the model conceptualisation by LeCun \citep{lecun2022path}.In this formulation,the learning of a dynamic environment is decomposed into three tasks: an encoder which compresses a high dimensional observation into a latent space, a memory which carries the internal state of the environment forward in time, and a predictor which maps the current state and input to the next observation. This decomposition can be mapped onto the components of a multi-scale material problem. The material architecture is the spatial observation to be encoded, the damage state is the environment state to be tracked over time, and the stress is the quantity to be predicted. In a multi-scale simulation problem, the role of the controller in a World Model, is assumed by the Finite Element Solver, which supplies the strain increment at each step, and is not learned.  This mapping aligns with the conventional phenomenological damage models for composites, where the material response is calculated from the applied strains and a set of internal state variables which track damage initiation and progression \citep{puck1998failure,ladeveze2000mesomodel}. In this sense, the memory block plays the role that the damage state variables play in a conventional constitutive law. A separate memory instance is required for every integration point in the structural model, and the internal state of each instance must be retained between solver increments. For a wind turbine blade or an aircraft wing, this can extend to millions of integration points. Consequently, the memory block is the component which determines whether the framework can be deployed at structural scale. Here, the size of the neuron internal state, rather than accuracy alone, becomes a governing design constraint. 

 In a multi-scale simulation the history is comprised of two components with different origins. The first is the strain, which is imposed by the FE solver as a result of the structural boundary conditions and the global equilibrium state. The strain is therefore a forcing term supplied to the environment and is not predicted by the model. The second is the stress, which is generated by the model and returned to the solver, where it is assembled into the internal force vector and consequently contributes indirectly to the strains applied at the following increment.  Here, a small error in the surrogate predicted stress can propagate into the strains subsequently imposed by the solver, and over a large number of increments this leads to the cumulative error drift previously observed for recurrent models of composite materials \citep{elsaid2023predicting}. To mitigate this error drift, a recursive training stage is employed where the composite model is trained using it is own stresses predictions as input. 

The proposed system (Figure~\ref{fig:mmst_arch}) is comprised of four main components:

\begin{figure}
\centering
\includegraphics[width=0.75\textwidth]{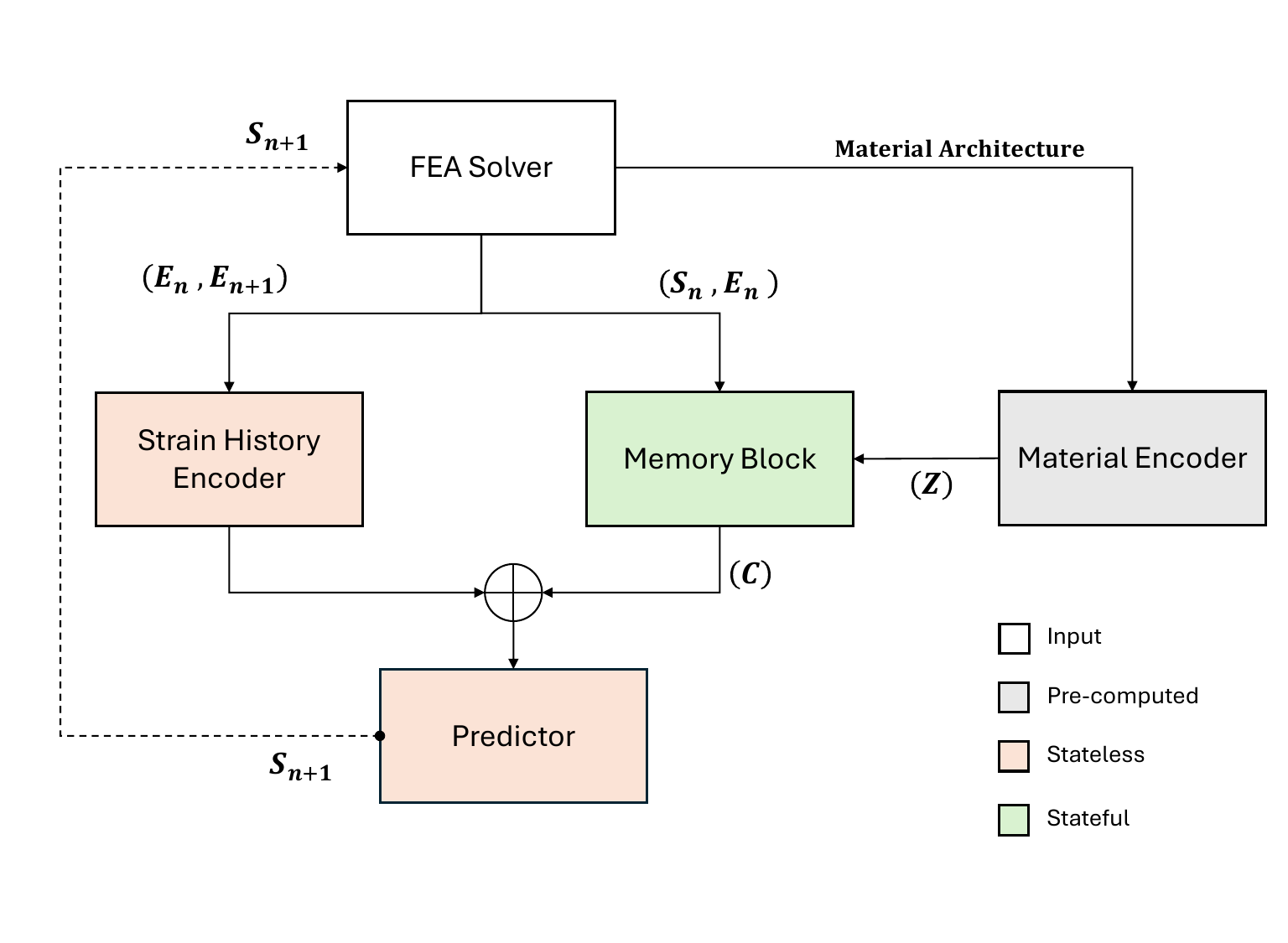}
\caption{Composites World Model: Multi-modal Transformer Architecture for Multi-scale
Modelling.}
\label{fig:mmst_arch}
\end{figure}
\paragraph{Strain History Encoder} Transformer block which encodes the short-term strain history, in addition to the strain at the current step calculated by the FE solver. In this model, time dependent inputs and outputs are tokenised with a token length of 3 time steps. This is the minimum token size that allows the transformer to see the status of loading/unloading at a given time step

\paragraph{Material Architecture Encoder} A Beta Variational Auto-Encoder ($\beta$-VAE) learns the material architecture including defects and variabilities and encodes it into a material latent space. This acts as a material embedding added to the predictor block context.This is analogues to the observation function in World Models.

\paragraph{Predictor} A transformer block which takes as input the encoded material state from the memory block and the encoded strain state and predicts the stresses for the next time step. The output from the predictor is fed back to the FE solver, which uses this prediction from each integration point to calculate the entire structure state for the next time-step.

\paragraph{Memory Blocks} The memory block generates the input context for the predictor block. It performs the function of tracking the material internal state and feeds it to the predictor block at each time step. This enables tracking long term information, but at the same time since the memory functions are limited to these blocks, they are the only part of the model that is required to be stateful and consequently need information tracking between each iteration.

The SNN implementation in the proposed model follows the simplified Leaky Integrate-and-Fire (LIF) formulation of
Eshraghian et al.\citep{eshraghian2023training}, which follows from the model by Neftci et al.\citep{neftci2019surrogate}.
The neuron is treated as a parallel resistor-capacitor circuit that allows a small portion of the capacitance to leak over time. This formulation is selected as it has the smallest number of hyper-parameters to optimise, and was shown to give similar performance to more complex versions developed to more closely replicate biological neurons. The input current to each neuron is given by,
\begin{equation}
\mathbf{I}_n = \mathbf{W}\,\mathbf{x}_n + \mathbf{U}\,\mathbf{S}_{n-1} + \mathbf{b},
\label{eq:current}
\end{equation}
where $\mathbf{x}_n$ is the input vector to the layer at time step $n$, $\mathbf{S}_{n-1}$ is the spike output of the same layer at the previous time step, and $\mathbf{W}$, $\mathbf{U}$ and $\mathbf{b}$ are the learned input weights, recurrent weights and biases respectively. Here, the membrane resistance is absorbed into the learned weights and biases. Consequently, the resistance is not treated explicitly as in a full LIF implementation. The membrane potential is then updated in discrete form as,
\begin{equation}
V_n = \delta\,V_{n-1} + (1-\delta)\,I_n,
\label{eq:lif}
\end{equation}
where $V_n$ and $V_{n-1}$ represent the neuron membrane potential
for the current and previous time steps respectively. The decay factor $\delta$ is given by,
\begin{equation}
\delta = e^{-1/\tau},
\label{eq:delta}
\end{equation}
where $\tau$ is the membrane time constant, a hyper-parameter
controlling how fast the neuron potential decays over time. A spike $S_n = H(V_n - V_{\mathrm{th}})$ is
emitted when the membrane potential reaches the threshold
$V_{\mathrm{th}}$, after which $V_{\mathrm{th}}$ is subtracted from $V_n$.
The derivative of the Heaviside function is replaced during back-propagation by a fast-sigmoid surrogate gradient\citep{neftci2019surrogate},
\begin{equation}
\frac{\partial S_n}{\partial V_n} \approx
\frac{1}{1 + k\,\lvert V_n - V_{\mathrm{th}} \rvert},
\label{eq:surrogate}
\end{equation}
where $k$ is the surrogate slope. In this work, $\tau = 10$, $V_{\mathrm{th}} = 1.0$ and $k = 10$ are used in both SNN layers.

Figure~\ref{fig:mmst_arch} shows a simplified representation of the proposed model. The input to the proposed system consistent of two data modes. Spatial information in the form of material architecture alongside time-dependent homogenised stress/strain data. In this paper, laminated composites will be used to demonstrate the system, however the architecture is not limited to a specific type of material.

\section{Multi-Modal Spiking Transformers Training}
\label{sec:training}

 To mitigate the cumulative error drift characteristic of auto-regressive material models, the temporal training of the proposed World Model is split into two stages: base training and recursive training. The base training is concerned with training the model against the ground truth over short sequences, followed by recursive training stage  which cover complete sequences to ensure the model's robustness against errors in its own predictions. 

\subsection{Base Training}
\label{sec:base_model}

\begin{figure}
\centering
\includegraphics[width=0.95\textwidth]{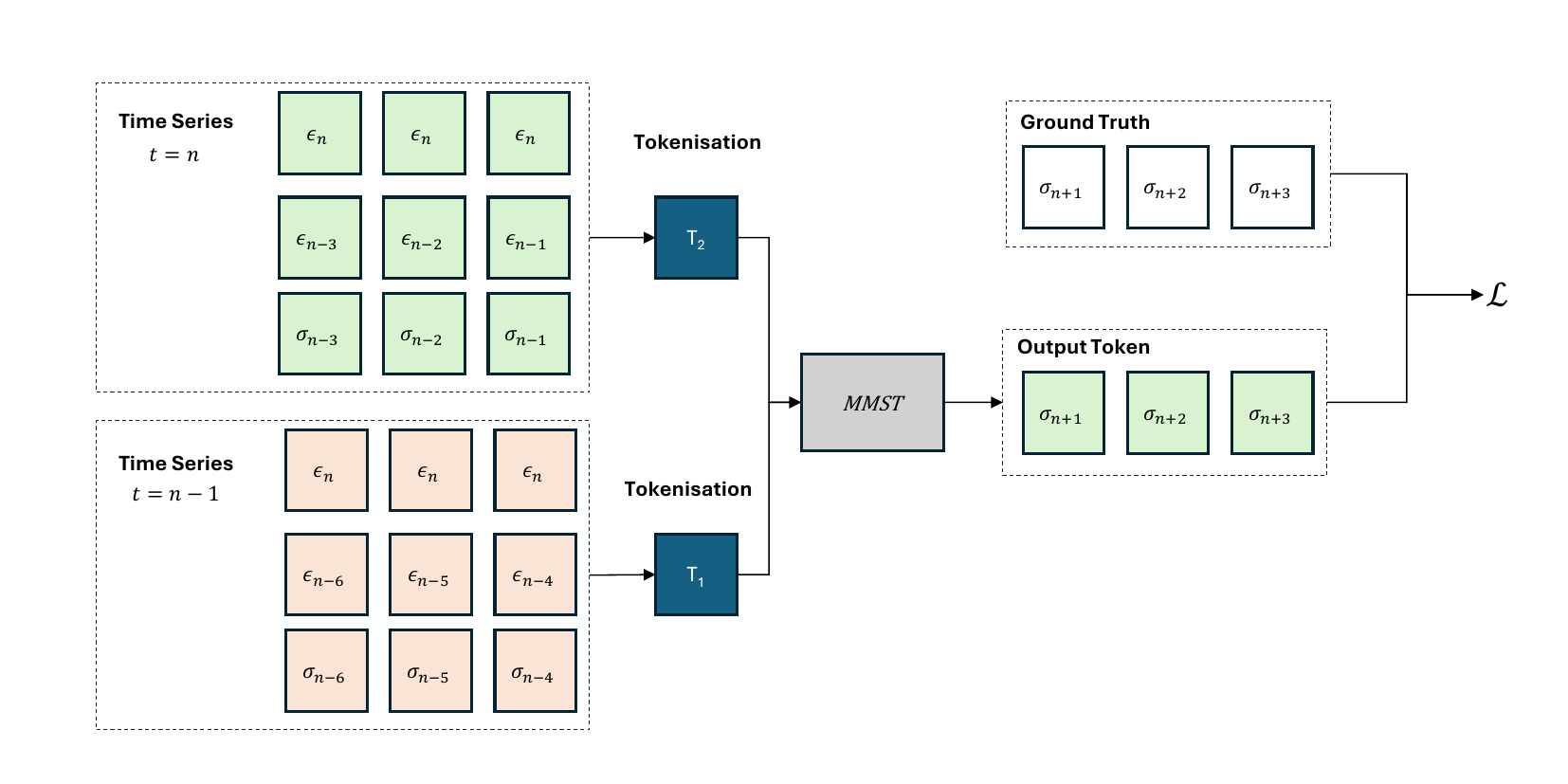}
\caption{Base Model Training Schematic, showing an MMST architecture with an input window (history) of 6 timestep and an output window of 3 timesteps (future).}
\label{fig:base_training}
\end{figure}

Base training, shown in Figure~\ref{fig:base_training}, allows the model to learn the short-term relationship between stress/strains and material architecture. The model is trained against the ground truth using supervised learning:

\begin{itemize}
\item All time series are segmented into output and input using a sliding window algorithm. These time series are strain history, stress history and the current strain. Both the window and input sizes are model parameters that can be chosen. Here, an input size of 6 time-steps (history), equivalent to two tokens and an output size of 3 time-step (1 token), are used.
\item The base model is trained on each time window separately, the input sequence is only reserved inside the time window.
\item The Loss function $\mathcal{L}$,  Equation~\eqref{eq:loss}, is calculated vs the ground truth for the full output token, 3 time-steps.
\end{itemize}

At this stage in training, the model has learned to predict the output for short sequences, which is insufficient for FE simulations, where the simulations can run for hundreds of thousands of time-steps for explicit time integration. The 2nd stage is concerned with training the model to predict in auto-regression mode, where the model uses its own stress predictions as part of its input. 

\subsection{Recursive Training}
\label{sec:recursive_model}

\begin{figure}
\centering
\includegraphics[width=0.85\textwidth]{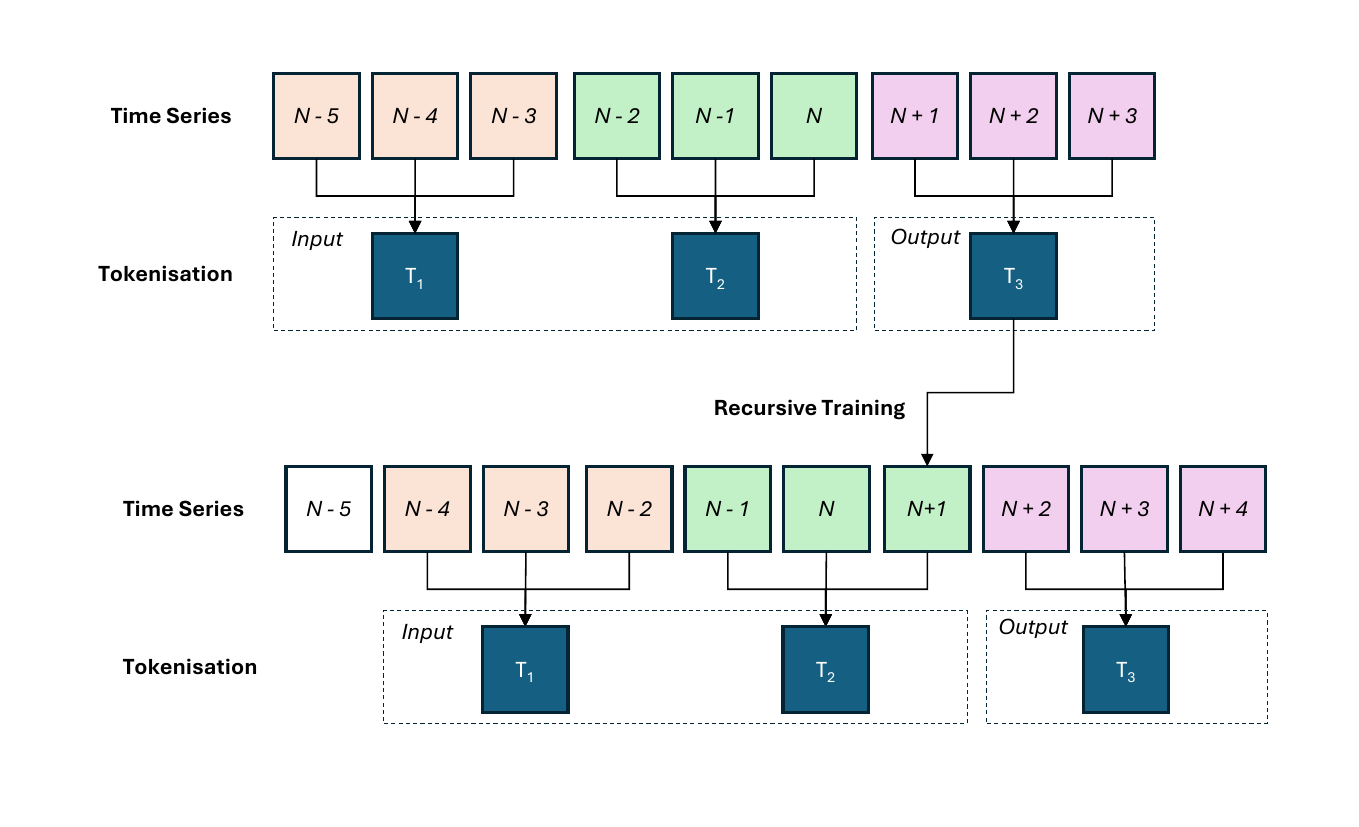}
\caption{Recursive training of proposed model, the stress output from the previous prediction is used as an input for the subsequent prediction.}
\label{fig:recursive_training}
\end{figure}

The recursive training approach is shown in Figure~\ref{fig:recursive_training}. For each time-step, an input window covering the stress and strains for the previous input window (6 time-steps) is submitted to the model. The model predicts a window of 3 time steps, which are then compared against the ground truth to calculate the Loss Function. For the next input window, the time-steps are shifted by 1 and the stress input is replaced by the model stress prediction from the previous timestep. This process is repeated for each time-step and for each sequence in a mini-batch, until the model has been trained on the entire dataset before the next epoch starts.

The recursive training focuses on making the model robust to errors in its own stress-predictions. Since, the previous stress predictions are input to the model the errors will accumulate over long predictions. Training solely against the ground-truth does not resolve this issue. The key differences introduced from the base-model
training are:
\begin{itemize}
\item  During recursive training, the transformer blocks ( Strain Encoder and Predictor), which has been already trained on the short-term response, are frozen and only SNN memory blocks are fine-tuned; to ensure that the model can converge during training over very long sequences:
\item The model is trained on each complete stress/strain sequence in each batch. While the entire database is submitted for training over a number of mini-batches, each mini-batch contains complete sequences ensuring the model learns the different stage of loading/unloading, damage initiation and progression.
\item The training employs a scheduled sampling approach. Here, early in training the input stresses are loaded directly from the RVE database. As training progresses, for an increasing fraction of the sequences, the ground truth stresses are replaced by the model own predictions. This means that for these recursive sequences, the model uses its previous stress predictions to predict the next time step. The recursion ratio increases each epoch until the model is fully trained using its own predicted stress. 
\end{itemize}

\subsection{Spatial Pre-training}
\label{sec:spatial}

The second input mode to the multi-modal transformer is the spatial
data representing the material architecture including defects and
variabilities. Each stress-strain sequence used for spatial training
is generated from a material-scale RVE model containing all necessary
information about the material architecture. For example, in the
composite material example shown later in this paper, the material
architecture is defined by the fibre orientation in each layer and
the presence of different defects such as wrinkles and matrix
pockets. An effective way to develop a relation between the material
architecture and their mechanical response, is to parametrise this
spatial information using an auto-encoder, which are deep neural
networks based on convolutional layers. In the proposed framework, a
De-noising $\beta$-Variational Auto-Encoder (D$\beta$-VAE) is
employed for this task. D$\beta$-VAE are an extension of the
standard Variational Auto-encoder that combines de-noising
training with $\beta$-regularised latent representation learning.

\begin{figure}
\centering
\includegraphics[width=0.85\textwidth]{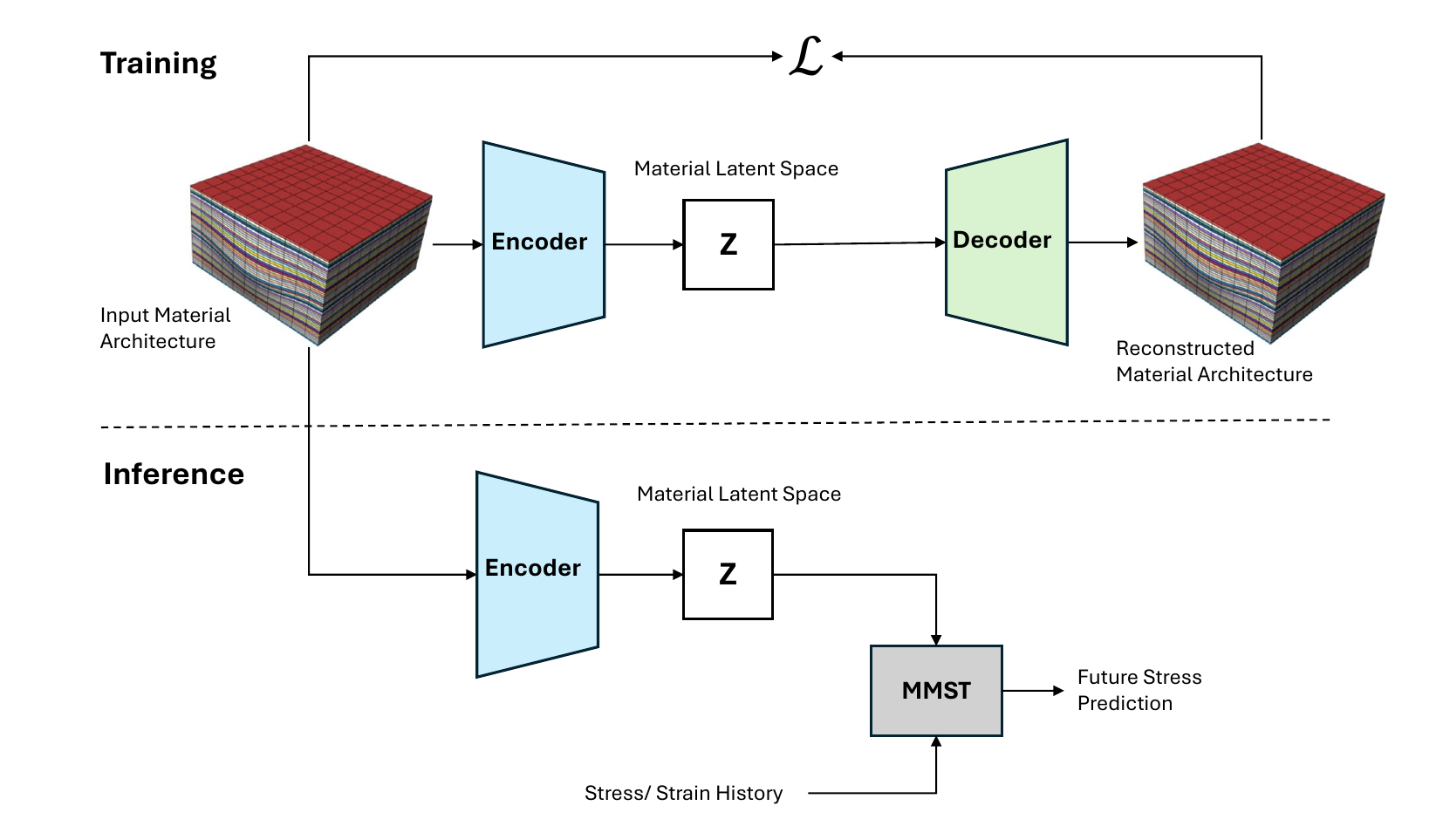}
\caption{De-noising--Beta Variational Autoencoder for Material Architecture Parametrisation, pre-training and integration with MMST during inference.}
\label{fig:dbvae}
\end{figure}

During training, input RVE images are deliberately corrupted with
Gaussian noise, and the encoder learns to infer a latent
distribution from this noisy input, while the decoder reconstructs
the original clean signal. This de-noising objective improves
robustness and forces the latent representation to capture essential
material features rather than noise. The $\beta$ parameter encourages
disentangled latent variables, promoting independence among latent dimensions.
As a result, D$\beta$-VAEs are ideal for creating a material latent
space where the different architectural features are encoded in a
continuous manner, and have seen wide use in material science applications \citep{Tao2021,Kissas2024}. 
In the proposed framework, a D$\beta$-VAE is pre-trained on the material architectures from the RVE database, 
where it learns to encode the material architecture which is used as input alongside the stress/ strain history to the MMST model.
This enables the rest of the deep learning framework to develop links between the material architecture and their mechanical response.
An overview of the D$\beta$-VAE pre-training and integration into the multi-modal system is shown in Figure~\ref{fig:dbvae}.

\section{Composites with Defects and Variability Database}
\label{sec:database}

\subsection{RVE Models}
\label{sec:rve_models}

\begin{figure}
\centering
\includegraphics[width=0.95\textwidth]{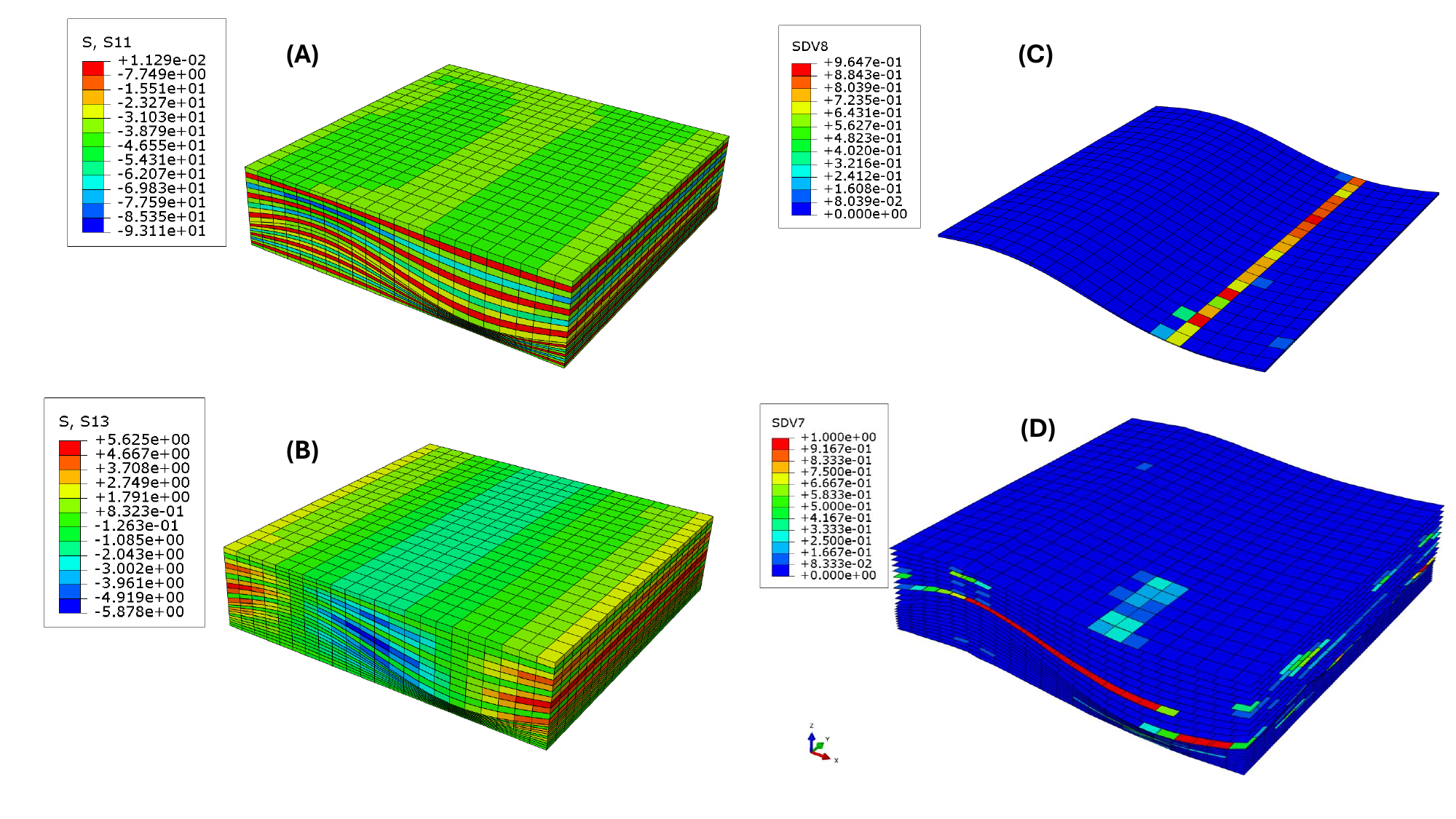}
\caption{Example composite RVE model with wrinkle defect. (A) Fibre direction stress distribution; (B) Transverse shear stress distribution; (C) Matrix cracking across a single layer; (D) Delamination progression through the wrinkle section.}
\label{fig:rve_example}
\end{figure}

Some of the most common defects in composites are wrinkles, gaps, matrix pockets with no fibres, and voids. These defects have been shown to have a significant adverse impact on the materials' mechanical
performance \citep{bender2019parametric,mehdikhani2019voids,mukhopadhyay2015tensile,mukhopadhyay2015compressive}.The proposed MMST framework have been trained on a database of
pre-computed RVE models of various composite layups under multiple loading conditions. For these models, the failure of a composite material is modelled on the meso-scale where the plies and interfaces are defined explicitly. Two separate models are used to predict matrix cracking and delamination failure modes respectively. For matrix cracking, a smeared crack model with energy regularisation is used following the work by Mukhopadhyay\citep{mukhopadhyay2015tensile,mukhopadhyay2015compressive}. To model delamination, the Cohesive Zone Model (CZM) proposed by Jiang\citep{jiang2007concise} is used. The RVE models are solved under periodic boundary conditions with the loading introduced as velocity control. Periodic homogenisation is a well-established technique for calculating
equivalent properties of composite materials \citep{li2008boundary,miehe2002computational}. Two types of defects are being studied in this work: wrinkles and gaps. Both defects are
modelled explicitly, where wrinkles are represented as a change in the layer geometries for RVEs affected by the defect (see Figure~\ref{fig:rve_example}), while the gap defect is
modelled as matrix regions within the layers. The RVE model dataset where generated using a commercial FEA solver (ABAQUS) combined with custom subroutines for composite damage initiation and progression. Example RVE is shown in Figure~\ref{fig:rve_example}.

In total 8400 unique RVE models were generated through a Monte-Carlo simulation. Each RVE model represents a unique combination of:
\begin{itemize}
\item \textit{Strain path}: the strain path is defined using an
initial 6 component strain state and a final 6 component strain
state, in addition to loading and unloading over the strain path.
This approach follows a database generation algorithm previously proposed by the authors \citep{elsaid2023predicting}.
\item \textit{Layup}: fibre orientation through the thickness is
generated randomly for each layer.
\item \textit{Defect and variabilities}: defect type, location and
severity are chosen randomly. A wrinkle is defined by a severity
angle ($0^{\circ}$--$11^{\circ}$). Gaps are defined by the location
and size.
\end{itemize}

All RVE models are solved using a non-linear FEA solver, then
submitted for computational homogenisation to generate full
stress/strain curves. The generated stress/strain series vary
between 200 time-steps and 1000 times steps in length. The complete
database consists of the set of strain
$(\varepsilon_{xx},\varepsilon_{yy},\varepsilon_{zz},
\varepsilon_{xz},\varepsilon_{zy},\varepsilon_{yx})$, stress curves
$(\sigma_{xx},\sigma_{yy},\sigma_{zz},
\sigma_{xz},\sigma_{zy},\sigma_{yx})$ and an RVE image representing
the layup and defects.

\subsection{Multi-Modal Spiking Transformer Training for Composites
with Defects Database}
\label{sec:training_composites}

A key concern when training the proposed system has been the model robustness to change in the size of the strain steps. FE solvers control the time step and consequently the strain step to maintain simulation stability. While the original database has been generated with different strain steps, it was found that the model achieves a more stable training through temporal data augmentation. This is achieved by resampling the time series to increase or decrease the strain step, to avoid the model learning the FEA time control behaviour instead of the impact of strain increments (Clever Hans effect). For training, the database has been resampled at 0.5, 0.75, 1.25 the original strain step. This results in 33{,}600 time-series for training. For base training,  the time series are windowed into overlapping time segments of 9 time-steps each, leading to 21M training points. Throughout this work, the RVE database is divided into 70\% for training and 30\% for validation.

Physics-based Machine Learning has become a staple of material response modelling using deep learning, which helps the model learn physically consistent behaviour and enables convergence with sparse datasets \citep{Polydoras2024, Xiong2024, Meyer2023, Ghosh2025}.In the proposed framework, this constraint is applied to the material stiffness tensor. The tensor $\boldsymbol{Q}$ eqn.\eqref{eq:stiffness} is obtained by differentiating the predicted stresses with respect to the applied strains. It has been shown that the 1OH homogenized stiffness tensor of a laminated composite follows a specific structure \citep{elsaid2018multiscale}. Enforcing this structure is a constraint on the pattern of the tensor rather than on its individual magnitudes. This is relevant here since the SNN backward pass uses a surrogate gradient. The magnitudes returned through the spiking layers are not guaranteed to be physically representative, but the symmetry of the tensor and the location of its zero terms are properties of the laminate rather than of the network and can be enforced directly.
\begin{equation}
\boldsymbol{Q} =
\begin{bmatrix}
Q_{11} & Q_{12} & Q_{13} & Q_{14} & 0      & 0      \\
       & Q_{22} & Q_{23} & Q_{24} & 0      & 0      \\
       &        & Q_{33} & Q_{34} & 0      & 0      \\
       &        &        & Q_{44} & 0      & 0      \\
       & \text{Sym} &    &        & Q_{55} & Q_{56} \\
       &        &        &        &        & Q_{66}
\end{bmatrix}
\label{eq:stiffness}
\end{equation}
This constraint has been enforced during training by penalising the
loss function when the network violates the symmetry conditions and
the non-zero terms. The loss function becomes:
\begin{equation}
\mathcal{L}_{\text{MMST}} = \mathcal{L}_{\text{STRESS}}
+ \lambda\,\mathcal{L}_{\text{SYM}}
+ \lambda\,\mathcal{L}_{\text{ZERO}}
\label{eq:loss}
\end{equation}
Where $\lambda$ is a scaling factor that adjusts the order of
magnitude on the stiffness loss terms to the same order of magnitude
as the stresses, set as 1E-3 for this work.

The proposed model is trained on the composites RVE database following the procedure outline in Section~\ref{sec:training}. A comparison of the memory block
performance is shown in Table~\ref{tab:memory_blocks}. Here, the strict sequential nature of SNN has proven to be the ideal choice for
modelling material non-linear response. SNN has to sequentially reach a certain membrane potential before the neuron starts spiking, which closely mimics the brittle fracture in composites and is behind its ability to capture the sudden drop in stresses as result of damage progression in composites. Since the main goal behind the introduction of SNN in this work has been memory efficiency, the comparison in Table~\ref{tab:memory_blocks} is conducted at a fixed memory block size. The block size has been chosen iteratively for SNN and later evaluated for LSTM and xLSTM. It should be noted that this is not a comparison of the three architectures at their individually optimal configurations. Even if all three perform at the same level of accuracy, SNN will be still the best choice for this work, given its lower memory budget, which is crucial for an FEA simulation as explained in Section~\ref{sec:dl_multiscale}
\begin{table}
\centering
\caption{Memory block type comparison following recursive training.}
\label{tab:memory_blocks}
\begin{tabular}{lccc}
\toprule
\textbf{Memory block} & \textbf{Block size} & \textbf{Epochs}
                     & \textbf{Validation error} \\
\midrule
SNN   & [128, 64] & 100 &  1{,}208 \\
xLSTM & [128, 64] & 100 & 49{,}731 \\
LSTM  & [128, 64] & 100 & 81{,}188 \\
\bottomrule
\end{tabular}
\end{table}

\begin{figure}
\centering
\includegraphics[width=0.95\textwidth]{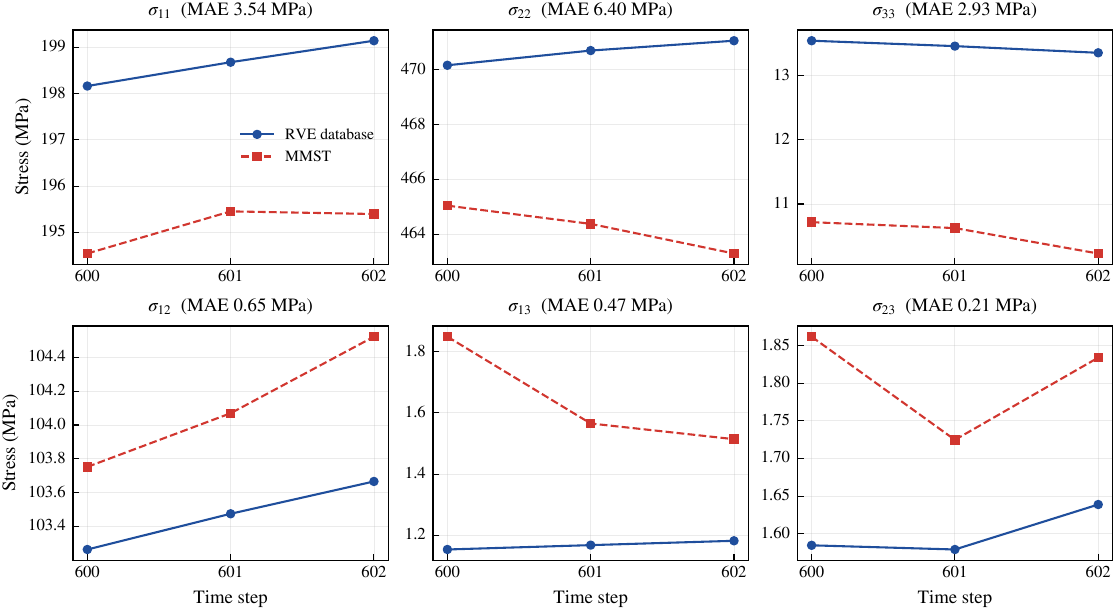}
\caption{Base Training output on a validation sample. The output is shown for all six stress components over a single output token (3-time Steps)}
\label{fig:base_sample}
\end{figure}

\begin{figure}
\centering
\includegraphics[width=0.75\textwidth]{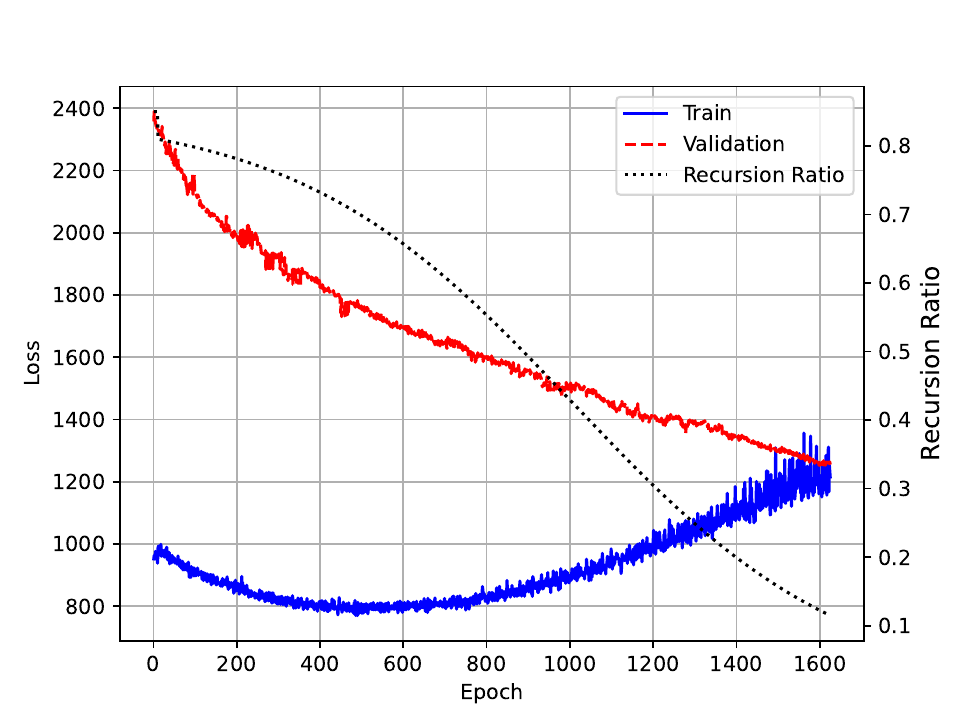} 
\caption{Recursive Training progression vs Epoch for validation and training sets. Validation is predicted fully in auto-regressive mode, while the training combines ground-truth input with auto-regression using the current recursion ratio for each epoch. } 
\label{fig:recursive_result}
\end{figure}

\begin{figure}
\centering
\includegraphics[width=0.95\textwidth]{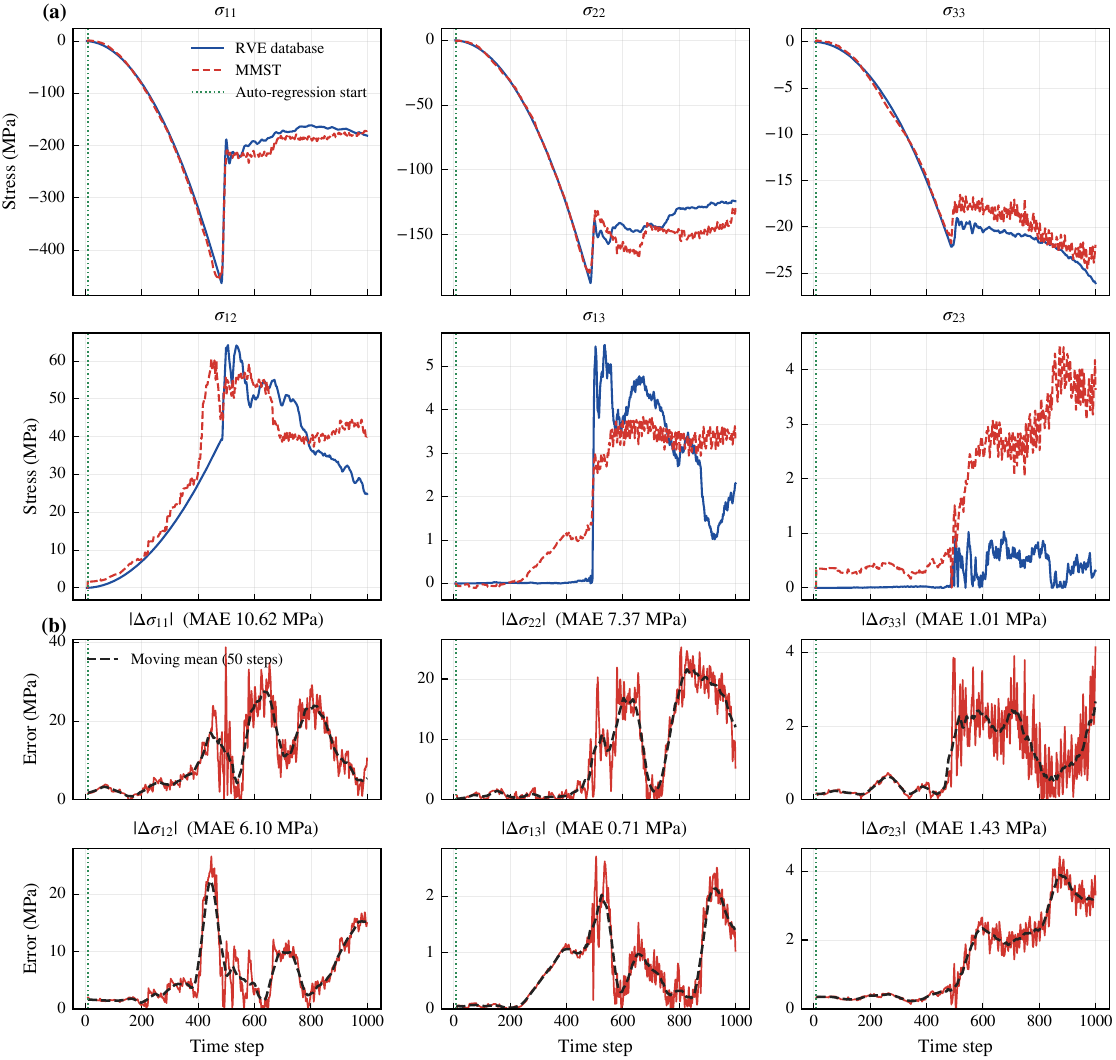} 
\caption{Example model prediction following recursive training on a validation sample for all stress components a) Stress predictions (MPa) vs RVE Database, b) Stress prediction error vs RVE Database (MPa)} 
\label{fig:mono_history}
\end{figure}

Sample outputs from the base training response are shown in Figure~\ref{fig:base_sample}. The evolution of the training error, validation error and recursion ratio during recursive training is shown in Figure~\ref{fig:recursive_result}a.  In the recrusive stage, the validation error is always calculated using full recursive prediction (auto-regression), where the model predicts the next stress step from the applied strains and its own previous stress prediction. Early in training, the training error reflects the high teacher forcing ratio at this stage, the model is driven by ground truth stresses, and by the end of training it is driven by its own outputs, matching the validation condition. 
The two curves illustrate why the recursive stage is required. Early in training the auto-regressive validation error is 2400, more than twice the training error of 1000, which is measured under teacher forcing and therefore under conditions the model will never encounter in during an FEA simulation. As the recursion ratio increases, the two converge to a common value of 1208, indicating that the model has been made robust to its own errors and can sustain accurate prediction from its own stress values. 
Figure~\ref{fig:mono_history} shows a full stress history prediction for a validation sample (chosen randomly) together with the associated error. The model reproduces the stress values in each direction, and the error envelope remains bounded over the full sequence rather than growing with the number of recursive steps.

\subsection{Memory Block Spiking Analysis}
\label{sec:spiking_analysis}

\begin{figure}
\centering
\includegraphics[width=0.8\textwidth]{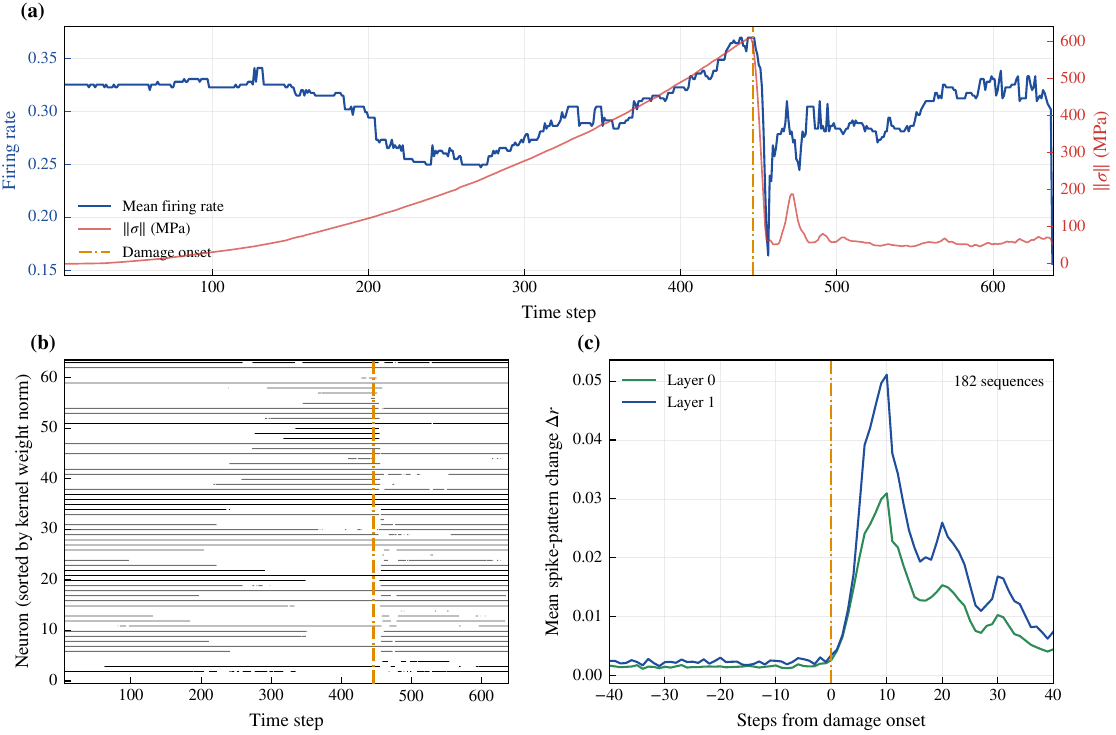}
\caption{SNN spiking patterns for an RVE under monotonic loading, transformer block input: A) mean SNN firing rate vs stress norm. B) SNN Neuron state vs time steps C) Mean change in SNN firing rate vs time step, for all Cross-Ply and UD samples, 182 samples }
\label{fig:mono_mean}
\end{figure}

A key challenge for surrogate modelling of materials undergoing damage progression is the ability to track the irreversible material internal state. This is essential for composites, where the accumulation of delamination, matrix cracking and fibre failure progressively reduces their load carrying capacity. Under monotonic loading, this internal state can be encoded implicitly, since the monotonically increasing load corresponds to a unique relation between the stresses and the total applied strain. Under loading and unloading this relation is no longer unique, and the same strain state can correspond to different stresses depending on the damage accumulated up to that point. Consequently, the internal state has to be tracked explicitly. In the proposed architecture, this task is assigned to the SNN memory block. This section is concerned with the SNN spiking patterns, and whether they demonstrate that the memory block is tracking damage initiation and progression. To this end, randomly chosen validation RVE models combining monotonic and cyclic loading where submitted for inference, with the neuron spikes recorded over time. The analysis showed a clear and irreversible switch in the spiking patterns once damage initiates. This behaviour is consistent between monotonic and cases with loading and unloading, demonstrating that the memory block is indeed successfully tracking the material state and not only correlating stress to strains. Example model prediction under monotonic loading is shown in Figure~\ref{fig:mono_history}. For this RVE, the damage initiates at time step 450. Comparing the neurons mean firing rate vs the stress norm, Figure~\ref{fig:mono_mean}a, shows a sharp drop in the neurons firing rate at the point of damage initiation from  0.36 to 0.16. Tracking the individual neuron's firing against loading (Figure~\ref{fig:mono_mean}b) shows that most of the neurons flip spiking patterns sharply as a result of damage initiation. The neurons in layer 0 and layer 1, which fire during elastic loading, stop firing following damage initiation and a new set of neurons take over. This strong correlation between SNN spiking patterns and the material state holds true for all tested samples. Figure~\ref{fig:mono_mean}c shows the change in SNN  firing rate $(\Delta r)$ in relation to the point of damage initiation, taken over 182 samples (Cross-Ply and UD), which further confirms the observation that the SNN learns the material state, and not only the strain.

\begin{figure}
\centering
\includegraphics[width=0.9\textwidth]{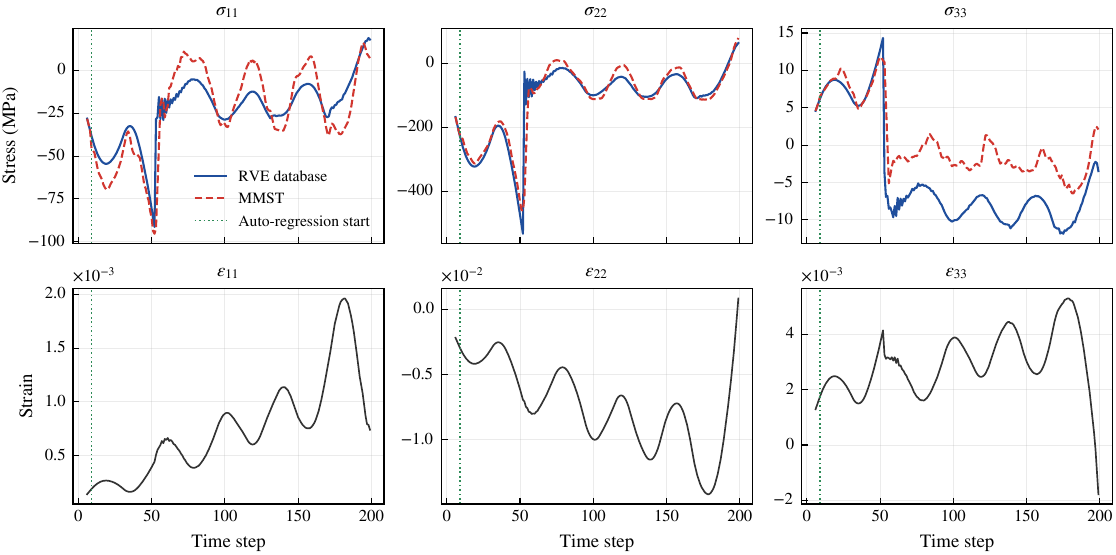}
\caption{Stress/strain history for an RVE under cyclic loading.}
\label{fig:cyc_history}
\end{figure}

\begin{figure}
\centering
\includegraphics[width=0.8\textwidth]{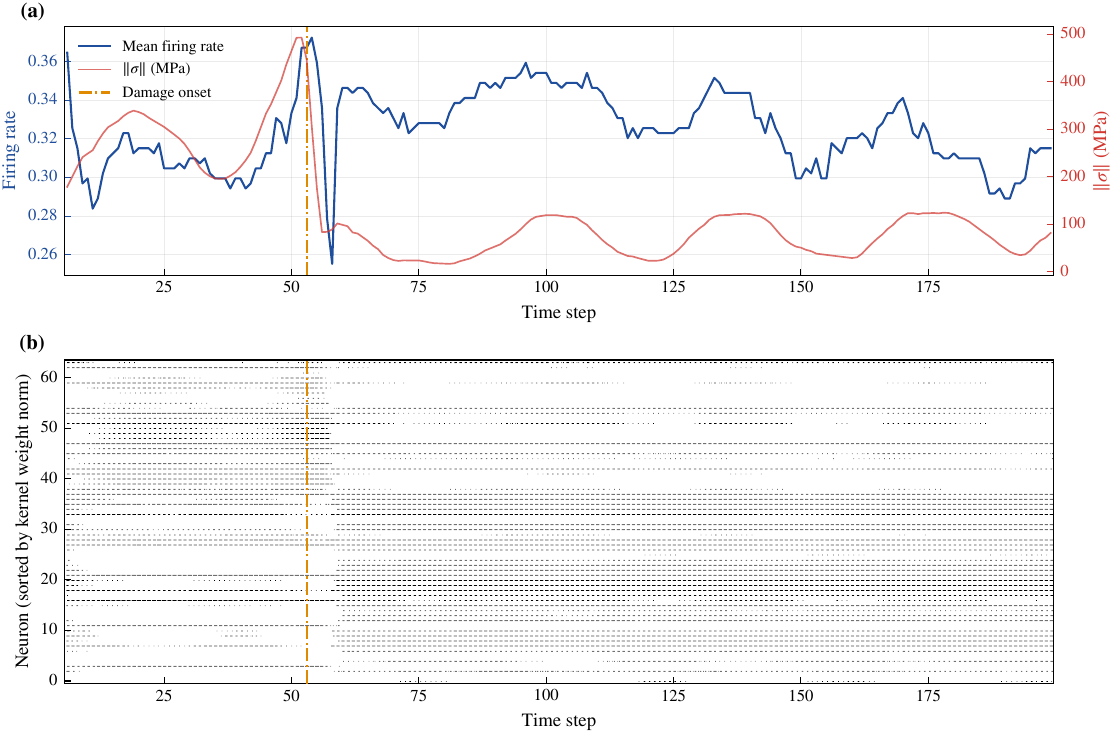}
\caption{SNN spiking patterns for an RVE under cyclic loading, transformer block input: A) mean SNN firing rate vs stress norm. B) SNN Neuron state vs time steps, the set of
neurons that fires during the elastic regime is largely silenced
once damage initiates and a different set takes over, after 53 steps.}
\label{fig:cyc_mean}
\end{figure}

A second example RVE under cyclic loading is shown in Figure~\ref{fig:cyc_history}. In this case, damage initiates at time step 53 followed by a rapid reduction in the material stiffness. The patterns seen for the monotonic loading case still holds under cyclic loading as shown in Figure~\ref{fig:cyc_mean}b. Additionally, the damaged spiking pattern continues to hold after damage initiation, even as the strain direction changes and the RVE is being unloaded (time step 53 - 200). This is a clear indication that the SNN block remembers and tracks the material damage state rather than encoding the stress in terms of strain only. Here, it is worth noting that this spiking behaviour have been observed for all tested RVE models and not only the two shown here.

\begin{figure}
\centering
\includegraphics[width=0.8\textwidth]{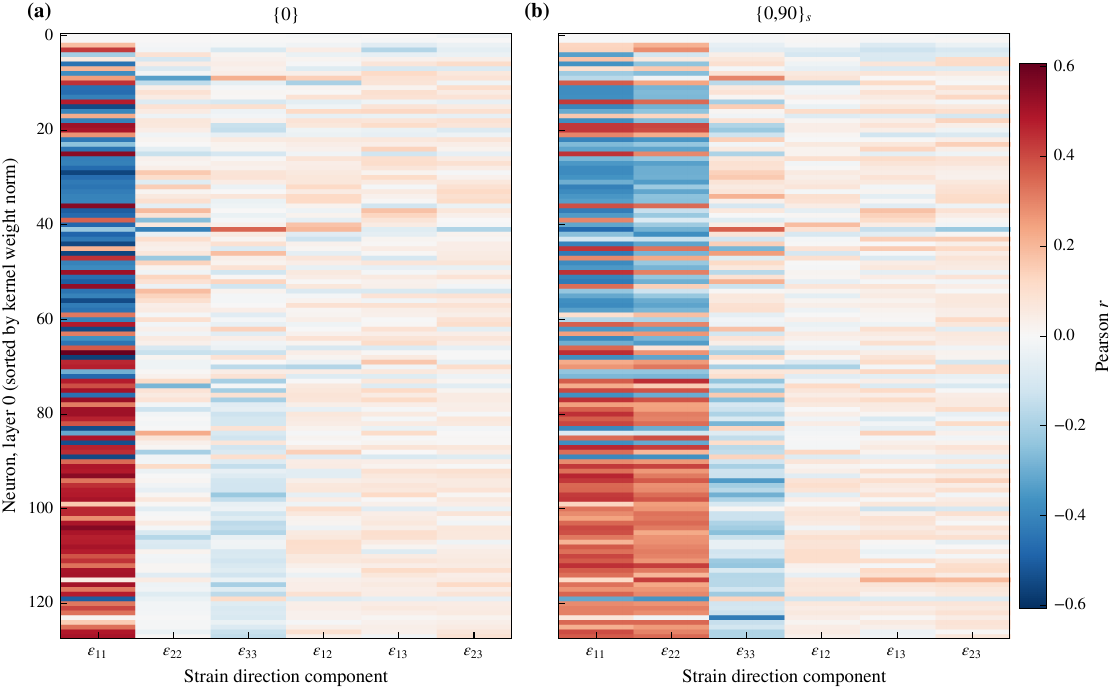}
\caption{Correlation between change in strain direction and SNN
neuron firing rate (Layer 0). Left: $0^{\circ}$ uni-directional
laminate. Right: $[0^{\circ},90^{\circ}]_S$ cross-ply laminate.}
\label{fig:correlation}
\end{figure}

Additionally, SNN firing rate shows strong correlation with the material architecture, visible in the change of neurons firing frequency in response to the change in fibre orientation. 
Here, two sets of 256 RVE models of a Uni-Directional (UD) composite [0] and a Cross-Ply (CP) [0, 90]$_S$ under various multi-axial loading conditions have been submitted
for inference with the SNN firing recorded. A Pearson correlation matrix between the change in the input strains and SNN firing rate is shown in Figure~\ref{fig:correlation}. 
In the case of the UD composite, where fibres are aligned along the $\varepsilon_1$ direction, the spiking has a strong correlation with the change in $\varepsilon_1$ . For the CP composite, where fibres are aligned along the both in-plane directions , the correlation expands to include changes in $\varepsilon_1$ and $\varepsilon_2$. This behaviour reflects the introduction of an orthogonal fibre direction in the CP composite, which leads to a stronger coupling between the two axial loading directions.  This behaviour indicates that the SNN is responding to the change in material architecture, as well as learning the material state.

\section{Finite Element Integration}
\label{sec:fe_integration}

The proposed Composites Wold Model has been developed
specifically to integrate with Finite Element solvers for
multi-scale simulations. Two main features have been introduced to
the architecture to guarantee an efficient and successful integration. The
model has been trained recursively to ensure it will be robust to the
cumulative error characteristics of time-dependent FE simulations.
The other feature is the modular nature of the model (Material Encoder,
Memory Block, Strain Encoder Block and Predictor Block). This was designed to enable model
disassembly for inference. While all the components of the
model has been trained concurrently, during inference it is
desirable to separate the components based on their sequence of
execution and memory requirements. At the start of an FE
simulation:
\begin{itemize}
\item The weights of the pre-trained MMST model are loaded and copied to
three separate DL models. These are the D$\beta$-VAE, Transformer
Blocks and the SNN Block. This effectively creates three models
that can be called independently.
\item The material architecture of all the FE integration points is
submitted to the D$\beta$-VAE and encoded into the material latent
space. This process is only executed once at the start of the
simulation, since the material architecture does not evolve during
the multi-scale simulation.
\item The FEA model being solved is submitted for domain decomposition based on the
maximum mini-batch size that the GPU can handle in one execution.
\item An SNN State Buffer is created to store the membrane potential
and spiking state of each neuron. This allows the SNN block to be
run as stateful and consequently be able to track the material
evolution for each integration point across the different FE solver
time steps.
\item The strains are calculated by the FE solver and then submitted
to the SNN block alongside the encoded material architecture. The
SNN will output the context information for the Transformer blocks.
\item The SNN output is submitted to the Transformer blocks to predict the stresses. Transformers use attention mechanisms and do not require memory to store a state.
\item The predicted stresses are fed back to
the FE solver to assemble the internal force vectors and progress
the simulation to the next time-step.
\end{itemize}

\begin{figure}
\centering
\includegraphics[width=0.9\textwidth]{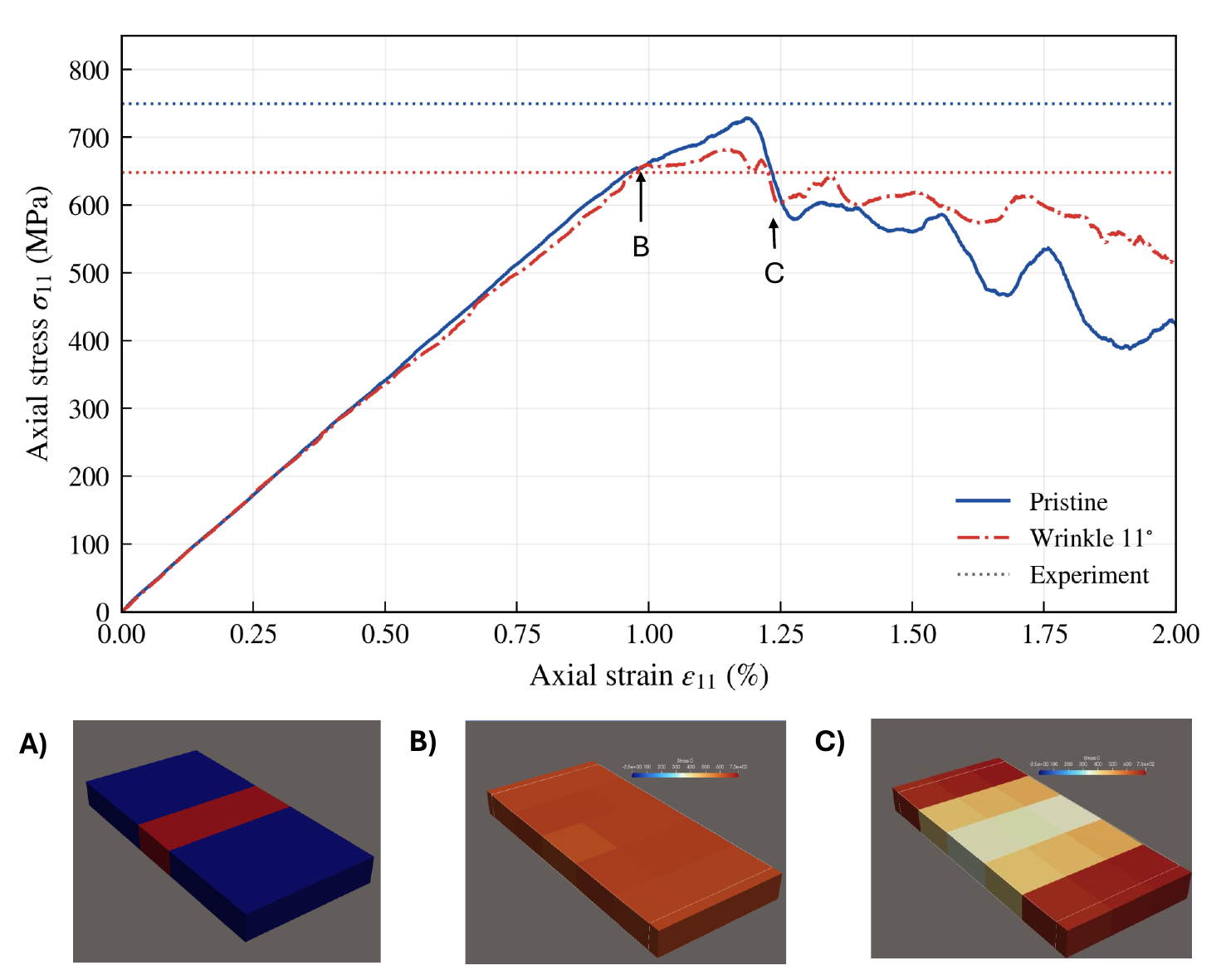}
\caption{Multi-scale FE model results for a QI specimen with embedded $11^{\circ}$ wrinkle, axial stress-strain curves for pristine/wrinkle under tension. A) FEA model setup showing region assigned a wrinkle RVE architecture, B) stress distribution before damage initiation, C) stress redistribution after damage}
\label{fig:Result_W_11_Tension}
\end{figure}

A schematic of the FE--MMST integration is shown in
Figure~\ref{fig:mmst_arch}. The Composite World Model has been integrated into an in-house multi-scale FEA solver, 7 multi-scale FEA modelling benchmarks were solved using the
integrated FE--MMST framework. These include composite
specimens with varying levels of internal defects under both compression and tension,
a summary of the results is shown in Table~\ref{tab:benchmark}. The
proposed system has been shown to be both accurate and reliable for
multi-scale simulation with errors ranging from 0.4\% to 4.8\%
against experimental results. The model is capable of generating
full history dependent homogenised stresses, while accounting
for material defects and layup. It can distinguish between pristine, high and low severity defects and
generate accurate material responses. The cumulative
error drift and the impractical memory requirements of the previous
generation of non-linear composite surrogate models has been 
mitigated. An example benchmark model is shown in Figure~\ref{fig:Result_W_11_Tension}.
The model was setup as a homogenised multi-scale simulation,
where each integration point is tracked by the MMST model. 
A defect region in the middle of the specimen is assigned a QI RVE with an embedded wrinkle of 11 degrees, while the rest of the specimen is assigned a pristine RVE. Loads are applied to the FEA model as prescribed displacement. The model shows very good agreement with the experimentally observed behaviour for this specimen. This includes damage initiation in the wrinkle region at 627 MPa (648 MPa Experimental). Damage initiation is followed by stress reduction in the wrinkle region and a drop in global stresses, a similar behaviour to what was reported experimentally.

\begin{table}
\centering
\caption{MMST benchmark vs experimental results for IM7/8552
composite material.}
\label{tab:benchmark}
\begin{tabular}{lcccc}
\toprule
\textbf{Benchmark} & \textbf{Defect severity} & \textbf{Prediction}
                  & \textbf{Experiment} & \textbf{Error} \\
\midrule
Tension - QI          & Pristine        & 753 & 750~\citep{mukhopadhyay2015tensile}     & 0.40\% \\
Compression - QI     & Pristine        & 627 & 643~\citep{mukhopadhyay2015compressive} & 2.49\% \\
Tension - QI          & Wrinkle $5^{\circ}$  & 750 & 717~\citep{mukhopadhyay2015tensile}     & 4.6\% \\
Compression - QI      & Wrinkle $5^{\circ}$  & 552 & 580~\citep{mukhopadhyay2015compressive}     & 4.83\% \\
Tension - QI          & Wrinkle $11^{\circ}$ & 657 & 648~\citep{mukhopadhyay2015tensile}     & 1.39\% \\
Compression- QI      & Wrinkle $11^{\circ}$ & 441 & 432~\citep{mukhopadhyay2015compressive}     & 2.08\% \\
Transverse Shear - CP & Pristine        &  79 &  83~\citep{tretiak2022predicting}       & 4.82\% \\
\bottomrule
\end{tabular}
\end{table}

\section{Conclusion}
\label{sec:discussion}

In this paper, a composite world model is integrated with a Spiking Neural Network Memory Block  for
modelling the non-linear behaviour of hierarchical materials has
been proposed. This novel model combines temporal and spatial
learning features to capture the impact of the material
architecture, defects and history dependencies. The model consists
of modular blocks that are trained jointly but are
disassembled during inference to allow for efficient model
execution during inference as part of a multi-scale modelling FE
simulation. The integrated FE--MMST model has been benchmarked
against experimental results for composites with embedded defects
and has shown good accuracy.

The SNN memory blocks used in this work generates context input for
the predictor block with the purpose of tracking the material internal
state, specifically damage evolution. The sequential nature of SNN
has been shown to be an ideal fit for this application.

The substantial performance gap between the SNN memory block and the equivalent
LSTM and xLSTM blocks, summarised in Table~\ref{tab:memory_blocks}, is consistent with the underlying SNN mechanism. 
The leaky integrate-and-fire dynamics impose a strictly sequential accumulation of membrane potential which mirrors
the monotonic accumulation of damage in the underlying RVE
response.  The lightweight per-neuron state additionally allows the
memory block to be deployed at every integration point of a
structural model without the memory expansion that has historically
limited LSTM-based surrogates from being applied at structural
scale.

The spiking analysis carried out for monotonically and
cyclically loaded RVE models showed a sharp and reproducible
reorganisation of the neuron firing patterns at the onset of damage
(Figures~\ref{fig:mono_mean},\ref{fig:cyc_mean}). The set of
neurons that fires during the elastic regime is largely silenced
once damage initiates, and a different set takes over. This
transition coincides with the change in the homogenised stress norm
and is consistent across loading paths and across RVE realisations.

Benchmarking against experimental results for IM7/8552 laminates
summarised in Table~\ref{tab:benchmark} show prediction errors
between 0.4\% and 4.8\% across the full set of configurations,
covering tension, compression and transverse shear, and pristine,
$5^{\circ}$ and $11^{\circ}$ wrinkle defect severities. The model
captures the relative strength reduction associated with increasing
wrinkle severity and reproduces the asymmetry between tensile and
compressive responses for the same defect. The absence of
cumulative error drift is attributed jointly to the recursive training stage with scheduled sampling and to the sequential dynamics of the SNN memory block.

\bibliography{references.bib}

@article{lomov2001textile,
  author  = {Lomov, S.V. and Huysmans, G. and Luo, Y. and Parnas, R.S.
             and Prodromou, A. and Verpoest, I. and Phelan, F.R.},
  title   = {Textile composites: modelling strategies},
  journal = {Composites Part A: Applied Science and Manufacturing},
  volume  = {32},
  number  = {10},
  pages   = {1379--1394},
  year    = {2001}
}

@article{desplentere2005microct,
  author  = {Desplentere, F. and Lomov, S.V. and Woerdeman, D.L. and
             Verpoest, I. and Wevers, M. and Bogdanovich, A.},
  title   = {Micro-{CT} characterization of variability in {3D}
             textile architecture},
  journal = {Composites Science and Technology},
  volume  = {65},
  number  = {13},
  pages   = {1920--1930},
  year    = {2005}
}

@article{selvaraj2023multiscale,
  author  = {Selvaraj, J. and El Said, B.},
  title   = {Multiscale modelling of strongly heterogeneous materials
             using geometry informed clustering},
  journal = {International Journal of Solids and Structures},
  volume  = {280},
  pages   = {112369},
  year    = {2023},
  doi     = {10.1016/j.ijsolstr.2023.112369}
}

@article{elsaid2016multiscale,
  author  = {El Said, B. and Ivanov, D. and Long, A.C. and Hallett, S.R.},
  title   = {Multi-scale modelling of strongly heterogeneous {3D}
             composite structures using spatial {V}oronoi tessellation},
  journal = {Journal of the Mechanics and Physics of Solids},
  volume  = {88},
  year    = {2016}
}

@article{hamila2008simulations,
  author  = {Hamila, N. and Boisse, P.},
  title   = {Simulations of textile composite reinforcement draping
             using a new semi-discrete three node finite element},
  journal = {Composites Part B: Engineering},
  volume  = {39},
  number  = {6},
  pages   = {999--1010},
  year    = {2008}
}

@article{mesogitis2014uncertainty,
  author  = {Mesogitis, T.S. and Skordos, A.A. and Long, A.C.},
  title   = {Uncertainty in the manufacturing of fibrous thermosetting
             composites: A review},
  journal = {Composites Part A: Applied Science and Manufacturing},
  volume  = {57},
  pages   = {67--75},
  year    = {2014}
}

@article{croft2011experimental,
  author  = {Croft, K. and Lessard, L. and Pasini, D. and Hojjati, M.
             and Chen, J. and Yousefpour, A.},
  title   = {Experimental study of the effect of automated fiber
             placement induced defects on performance of composite
             laminates},
  journal = {Composites Part A: Applied Science and Manufacturing},
  volume  = {42},
  number  = {5},
  pages   = {484--491},
  year    = {2011}
}

@article{heinecke2019manufacturing,
  author  = {Heinecke, F. and Willberg, C.},
  title   = {Manufacturing-Induced Imperfections in Composite Parts
             Manufactured via Automated Fiber Placement},
  journal = {Journal of Composites Science},
  volume  = {3},
  number  = {2},
  pages   = {56},
  year    = {2019}
}

@book{talreja2024failure,
  author    = {Talreja, R.},
  title     = {Failure analysis of composite materials with manufacturing
               defects},
  publisher = {CRC Press},
  year      = {2024}
}

@article{wisnom1996size,
  author  = {Wisnom, M.R. and Jones, M.I.},
  title   = {Size effects in interlaminar tensile and shear strength
             of unidirectional glass fibre/epoxy},
  journal = {Journal of Reinforced Plastics and Composites},
  volume  = {15},
  number  = {1},
  pages   = {2--15},
  year    = {1996}
}

@article{obert2014micro,
  author  = {Obert, E. and Daghia, F. and Lad{\`e}v{\`e}ze, P. and
             Ballere, L.},
  title   = {Micro and meso modeling of woven composites: Transverse
             cracking kinetics and homogenization},
  journal = {Composite Structures},
  volume  = {117},
  pages   = {212--221},
  year    = {2014}
}

@article{elsaid2018iterative,
  author  = {El Said, B. and Daghia, F. and Ivanov, D. and Hallett, S.R.},
  title   = {An iterative multiscale modelling approach for nonlinear
             analysis of {3D} composites},
  journal = {International Journal of Solids and Structures},
  volume  = {132--133},
  pages   = {42--58},
  year    = {2018}
}

@techreport{harris2003advanced,
  author      = {Harris, C.},
  title       = {Advanced Durability and Damage Tolerance Design and
                 Analysis Methods for Composite Structures: Lessons
                 Learned from {NASA} Technology Development},
  year        = {2003},
  institution = {NASA}
}

@article{pham2019discrete,
  author  = {Pham, D.C. and Cui, X. and Ren, X. and Lua, J.},
  title   = {A discrete crack informed {3D} continuum damage model and
             its application for delamination migration in composite
             laminates},
  journal = {Composites Part B: Engineering},
  volume  = {165},
  pages   = {554--562},
  year    = {2019}
}

@article{kashtalyan2013predicting,
  author  = {Kashtalyan, M. and Soutis, C.},
  title   = {Predicting residual stiffness of cracked composite laminates
             subjected to multi-axial inplane loading},
  journal = {Journal of Composite Materials},
  volume  = {47},
  number  = {20-21},
  pages   = {2513--2524},
  year    = {2013}
}

@article{puck1998failure,
  author  = {Puck, A. and Sch{\"u}rmann, H.},
  title   = {Failure analysis of {FRP} laminates by means of physically
             based phenomenological models},
  journal = {Composites Science and Technology},
  volume  = {58},
  number  = {7},
  pages   = {1045--1067},
  year    = {1998}
}

@article{ladeveze2000mesomodel,
  author  = {Lad{\`e}v{\`e}ze, P. and Allix, O. and De{\"u}, J.F. and
             L{\'e}v{\^e}que, D.},
  title   = {A mesomodel for localisation and damage computation in
             laminates},
  journal = {Computer Methods in Applied Mechanics and Engineering},
  volume  = {183},
  number  = {1-2},
  pages   = {105--122},
  year    = {2000}
}

@article{feyel2000fe2,
  author  = {Feyel, F. and Chaboche, J.L.},
  title   = {{FE}$^2$ multiscale approach for modelling the
             elastoviscoplastic behaviour of long fibre {SiC/Ti}
             composite materials},
  journal = {Computer Methods in Applied Mechanics and Engineering},
  volume  = {183},
  number  = {3-4},
  pages   = {309--330},
  year    = {2000}
}

@article{kouznetsova2002multi,
  author  = {Kouznetsova, V. and Geers, M.G.D. and Brekelmans, W.A.M.},
  title   = {Multi-scale constitutive modelling of heterogeneous
             materials with a gradient-enhanced computational
             homogenization scheme},
  journal = {International Journal for Numerical Methods in Engineering},
  volume  = {54},
  number  = {8},
  pages   = {1235--1260},
  year    = {2002}
}

@article{bessa2017framework,
  author  = {Bessa, M.A. and Bostanabad, R. and Liu, Z. and Hu, A. and
             Apley, D.W. and Brinson, C. and others},
  title   = {A framework for data-driven analysis of materials under
             uncertainty: Countering the curse of dimensionality},
  journal = {Computer Methods in Applied Mechanics and Engineering},
  volume  = {320},
  pages   = {633--667},
  year    = {2017}
}

@article{fritzen2016finite,
  author  = {Fritzen, F. and Hodapp, M.},
  title   = {The finite element square reduced ({FE}$^2$R) method with
             {GPU} acceleration: towards three-dimensional two-scale
             simulations},
  journal = {International Journal for Numerical Methods in Engineering},
  volume  = {107},
  number  = {10},
  pages   = {853--881},
  year    = {2016}
}

@article{kim2021deep,
  author  = {Kim, Y. and Kim, Y. and Yang, C. and Park, K. and Gu, G.X.
             and Ryu, S.},
  title   = {Deep learning framework for material design space
             exploration using active transfer learning and data
             augmentation},
  journal = {npj Computational Materials},
  volume  = {7},
  number  = {1},
  pages   = {140},
  year    = {2021}
}

@article{lim2021extrapolative,
  author  = {Lim, Y.F. and Ng, C.K. and Vaitesswar, U.S. and
             Hippalgaonkar, K.},
  title   = {Extrapolative Bayesian Optimization with {G}aussian Process
             and Neural Network Ensemble Surrogate Models},
  journal = {Advanced Intelligent Systems},
  volume  = {3},
  number  = {11},
  pages   = {2100101},
  year    = {2021}
}

@misc{ghane2023recurrent,
  author       = {Ghane, E. and Fagerstr{\"o}m, M. and Mirkhalaf, M.},
  title        = {Recurrent neural networks and transfer learning for
                  elasto-plasticity in woven composites},
  year         = {2023},
  howpublished = {arXiv:2311.13434},
  eprint       = {2311.13434},
  archivePrefix= {arXiv}
}

@article{kovacs2025physically,
  author  = {Kov{\'a}cs, N. and Maia, M.A. and Rocha, I.B.C.M. and
             Furtado, C. and Camanho, P.P. and van der Meer, F.P.},
  title   = {Physically Recurrent Neural Networks for computational
             homogenization of composite materials with microscale
             debonding},
  journal = {European Journal of Mechanics A/Solids},
  volume  = {112},
  pages   = {105668},
  year    = {2025},
  doi     = {10.1016/j.euromechsol.2025.105668}
}

@article{hii2022kinematically,
  author  = {Hii, A.K.W. and El Said, B.},
  title   = {A kinematically consistent second-order computational
             homogenisation framework for thick shell models},
  journal = {Computer Methods in Applied Mechanics and Engineering},
  volume  = {398},
  pages   = {115136},
  year    = {2022}
}

@article{neftci2019surrogate,
  author  = {Neftci, E.O. and Mostafa, H. and Zenke, F.},
  title   = {Surrogate Gradient Learning in Spiking Neural Networks},
  journal = {IEEE Signal Processing Magazine},
  volume  = {36},
  number  = {6},
  pages   = {51--63},
  year    = {2019}
}

@article{elsaid2023predicting,
  author  = {El Said, B.},
  title   = {Predicting the non-linear response of composite materials
             using deep recurrent convolutional neural networks},
  journal = {International Journal of Solids and Structures},
  volume  = {276},
  pages   = {112334},
  year    = {2023},
  doi     = {10.1016/j.ijsolstr.2023.112334}
}

@article{bender2019parametric,
  author  = {Bender, J.J. and Hallett, S.R. and Lindgaard, E.},
  title   = {Parametric study of the effect of wrinkle features on the
             strength of a tapered wind turbine blade sub-structure},
  journal = {Composite Structures},
  volume  = {218},
  pages   = {120--129},
  year    = {2019}
}

@article{mehdikhani2019voids,
  author  = {Mehdikhani, M. and Gorbatikh, L. and Verpoest, I. and
             Lomov, S.V.},
  title   = {Voids in fiber-reinforced polymer composites: A review on
             their formation, characteristics, and effects on mechanical
             performance},
  journal = {Journal of Composite Materials},
  volume  = {53},
  number  = {12},
  pages   = {1579--1669},
  year    = {2019}
}

@article{mukhopadhyay2015tensile,
  author  = {Mukhopadhyay, S. and Jones, M.I. and Hallett, S.R.},
  title   = {Tensile failure of laminates containing an embedded
             wrinkle; numerical and experimental study},
  journal = {Composites Part A: Applied Science and Manufacturing},
  volume  = {77},
  pages   = {219--228},
  year    = {2015}
}

@article{mukhopadhyay2015compressive,
  author  = {Mukhopadhyay, S. and Jones, M.I. and Hallett, S.R.},
  title   = {Compressive failure of laminates containing an embedded
             wrinkle; experimental and numerical study},
  journal = {Composites Part A: Applied Science and Manufacturing},
  volume  = {73},
  pages   = {132--142},
  year    = {2015}
}

@article{jiang2007concise,
  author  = {Jiang, W. and Hallett, S.R. and Green, B.G. and Wisnom, M.R.},
  title   = {A concise interface constitutive law for analysis of
             delamination and splitting in composite materials and its
             application to scaled notched tensile specimens},
  journal = {International Journal for Numerical Methods in Engineering},
  volume  = {69},
  number  = {9},
  pages   = {1982--1995},
  year    = {2007}
}

@article{li2008boundary,
  author  = {Li, S.},
  title   = {Boundary conditions for unit cells from periodic
             microstructures and their implications},
  journal = {Composites Science and Technology},
  volume  = {68},
  number  = {9},
  pages   = {1962--1974},
  year    = {2008}
}

@article{miehe2002computational,
  author  = {Miehe, C. and Schr{\"o}der, J. and Becker, M.},
  title   = {Computational homogenization analysis in finite elasticity:
             material and structural instabilities on the micro- and
             macro-scales of periodic composites and their interaction},
  journal = {Computer Methods in Applied Mechanics and Engineering},
  volume  = {191},
  number  = {44},
  pages   = {4971--5005},
  year    = {2002}
}

@article{elsaid2018multiscale,
  author  = {El Said, B. and Hallett, S.R.},
  title   = {Multiscale surrogate modelling of the elastic response of
             thick composite structures with embedded defects and
             features},
  journal = {Composite Structures},
  volume  = {200},
  pages   = {781--798},
  year    = {2018}
}

@article{tretiak2022predicting,
  author  = {Tretiak, I. and Kawashita, L.F. and Hallett, S.R.},
  title   = {Predicting short beam shear strength reduction in
             carbon/epoxy laminates containing voids},
  journal = {Composite Structures},
  volume  = {290},
  pages   = {115472},
  year    = {2022}
}

@article{ElFallakiIdrissi2024,
	author = {El Fallaki Idrissi, M. and Praud, F. and Meraghni, F. and Chinesta, F. and Chatzigeorgiou, G.},
	doi = {10.1016/j.jmps.2024.105604},
	issn = {00225096},
	journal = {Journal of the Mechanics and Physics of Solids},
	number = {March},
	pages = {105604},
	publisher = {Elsevier Ltd},
	title = {{Multiscale Thermodynamics-Informed Neural Networks (MuTINN) towards fast and frugal inelastic computation of woven composite structures}},
	url = {https://doi.org/10.1016/j.jmps.2024.105604},
	volume = {186},
	year = {2024}
}

@article{Ghane2024,
archivePrefix = {arXiv},
arxivId = {arXiv:2311.13434},
author = {Ghane, Ehsan and Fagerstr{\"{o}}m, Martin and Mirkhalaf, Mohsen},
doi = {10.1016/j.euromechsol.2024.105378},
eprint = {arXiv:2311.13434},
issn = {09977538},
journal = {European Journal of Mechanics, A/Solids},
title = {{Recurrent neural networks and transfer learning for predicting elasto-plasticity in woven composites}},
volume = {107},
year = {2024}
}

@article{Maia2023,
archivePrefix = {arXiv},
arxivId = {2209.07320},
author = {Maia, M. A. and Rocha, I. B.C.M. and Kerfriden, P. and van der Meer, F. P.},
doi = {10.1016/J.CMA.2023.115934},
eprint = {2209.07320},
issn = {0045-7825},
journal = {Computer Methods in Applied Mechanics and Engineering},
month = {mar},
pages = {115934},
publisher = {North-Holland},
title = {{Physically recurrent neural networks for path-dependent heterogeneous materials: Embedding constitutive models in a data-driven surrogate}},
url = {https://www.sciencedirect.com/science/article/pii/S0045782523000579},
volume = {407},
year = {2023}
}

@article{Methods2024,
author = {Methods, Comput and Mech, Appl and Zhongbo, Yuan and Hien, Poh Leong},
doi = {10.1016/j.cma.2024.116745},
issn = {0045-7825},
journal = {Computer Methods in Applied Mechanics and Engineering},
number = {August 2023},
pages = {116745},
publisher = {Elsevier B.V.},
title = {{Pre-trained transformer model as a surrogate in multiscale computational homogenization framework for elastoplastic composite materials subjected to generic loading paths}},
url = {https://doi.org/10.1016/j.cma.2024.116745},
volume = {421},
year = {2024}
}

@article{Rane2023,
author = {Rane, Nitin},
doi = {10.2139/ssrn.4609920},
journal = {SSRN Electronic Journal},
title = {{Transformers in Material Science: Roles, Challenges, and Future Scope}},
year = {2023}
}

@article{Tao2021,
author = {Tao, Chongcong and Zhang, Chao and Ji, Hongli and Qiu, Jinhao},
doi = {10.1016/j.compscitech.2020.108573},
issn = {02663538},
journal = {Composites Science and Technology},
number = {August 2020},
pages = {108573},
publisher = {Elsevier Ltd},
title = {{Application of neural network to model stiffness degradation for composite laminates under cyclic loadings}},
url = {https://doi.org/10.1016/j.compscitech.2020.108573},
volume = {203},
year = {2021}
}

@article{Kissas2024,
archivePrefix = {arXiv},
arxivId = {2402.04263},
author = {Kissas, Georgios and Mishra, Siddhartha and Chatzi, Eleni and {De Lorenzis}, Laura},
eprint = {2402.04263},
pages = {1--43},
title = {{The Language of Hyperelastic Materials}},
url = {http://arxiv.org/abs/2402.04263},
year = {2024}
}

@article{Xiong2024,
author = {Xiong, Zhihao and Yang, Ping and Zhao, Pengyang},
doi = {10.1016/j.ijmecsci.2024.109214},
issn = {00207403},
journal = {International Journal of Mechanical Sciences},
number = {October 2023},
pages = {109214},
publisher = {Elsevier Ltd},
title = {{Physics-driven neural networks for nonlinear micromechanics}},
url = {https://doi.org/10.1016/j.ijmecsci.2024.109214},
volume = {273},
year = {2024}
}

@article{Polydoras2024,
author = {Polydoras, Vasileios and Tandale, Saurabh and Stoffel, Marcus},
doi = {10.1002/PAMM.202400066},
issn = {1617-7061},
journal = {PAMM},
month = {dec},
number = {4},
publisher = {Wiley},
title = {{Physics‐Based Spiking Neural Network as a Surrogate Model for Viscoplastic Material Law in Impulsively Loaded Beams}},
volume = {24},
year = {2024}
}

@article{Meyer2023,
author = {Meyer, Knut Andreas and Ekre, Fredrik},
doi = {10.1016/j.jmps.2023.105416},
issn = {00225096},
journal = {Journal of the Mechanics and Physics of Solids},
number = {August},
pages = {105416},
publisher = {Elsevier Ltd},
title = {{Thermodynamically consistent neural network plasticity modeling and discovery of evolution laws}},
url = {https://doi.org/10.1016/j.jmps.2023.105416},
volume = {180},
year = {2023}
}

@article{Ghosh2025,
author = {Ghosh, Gourab and Biswas, Dhiraj and Bhattacharyya, Rudraprasad},
doi = {10.1016/j.compositesb.2025.112819},
issn = {13598368},
journal = {Composites Part B: Engineering},
number = {June},
pages = {112819},
publisher = {Elsevier Ltd},
title = {{Advancements in multiscale modeling of damage in composite materials: A comprehensive review}},
url = {https://doi.org/10.1016/j.compositesb.2025.112819},
volume = {307},
year = {2025}
}

@inproceedings{bulatov2022recurrent,
  author    = {Bulatov, Aydar and Kuratov, Yury and Burtsev, Mikhail S.},
  title     = {Recurrent Memory Transformer},
  booktitle = {Advances in Neural Information Processing Systems},
  volume    = {35},
  pages     = {11079--11091},
  year      = {2022}
}

@article{munkhdalai2024leave,
  author        = {Munkhdalai, Tsendsuren and Faruqui, Manaal and Gopal, Siddharth},
  title         = {Leave No Context Behind: Efficient Infinite Context Transformers
                   with Infini-attention},
  journal       = {arXiv preprint arXiv:2404.07143},
  year          = {2024},
  eprint        = {2404.07143},
  archivePrefix = {arXiv},
  primaryClass  = {cs.CL}
}

@article{ha2018world,
  author        = {Ha, David and Schmidhuber, J{\"u}rgen},
  title         = {World Models},
  journal       = {arXiv preprint arXiv:1803.10122},
  year          = {2018},
  eprint        = {1803.10122},
  archivePrefix = {arXiv},
  primaryClass  = {cs.LG},
  doi           = {10.5281/zenodo.1207631}
}

@inproceedings{hafner2019learning,
  author    = {Hafner, Danijar and Lillicrap, Timothy and Fischer, Ian and
               Villegas, Ruben and Ha, David and Lee, Honglak and Davidson, James},
  title     = {Learning Latent Dynamics for Planning from Pixels},
  booktitle = {Proceedings of the 36th International Conference on Machine Learning},
  series    = {Proceedings of Machine Learning Research},
  volume    = {97},
  pages     = {2555--2565},
  year      = {2019},
  publisher = {PMLR}
}

@article{ghavamian2019accelerating,
  author  = {Ghavamian, F. and Simone, A.},
  title   = {Accelerating multiscale finite element simulations of
             history-dependent materials using a recurrent neural network},
  journal = {Computer Methods in Applied Mechanics and Engineering},
  volume  = {357},
  pages   = {112594},
  year    = {2019},
  doi     = {10.1016/j.cma.2019.112594}
}

@article{eshraghian2023training,
  author  = {Eshraghian, Jason K. and Ward, Max and Neftci, Emre O. and Wang, Xinxin and Lenz, Gregor and Dwivedi, Girish and Bennamoun, Mohammed and Jeong, Doo Seok and Lu, Wei D.},
  title   = {Training Spiking Neural Networks Using Lessons From Deep Learning},
  journal = {Proceedings of the IEEE},
  volume  = {111},
  number  = {9},
  pages   = {1016--1054},
  year    = {2023},
  doi     = {10.1109/JPROC.2023.3308088}
}

@article{lecun2022path,
  title={A path towards autonomous machine intelligence version 0.9. 2, 2022-06-27},
  author={LeCun, Yann and others},
  journal={Open Review},
  volume={62},
  number={1},
  pages={1--62},
  year={2022}
}
\appendix
\section{Layer(0) Spiking Patterns}
\begin{figure}[h]
\centering
\includegraphics[width=0.75\textwidth]{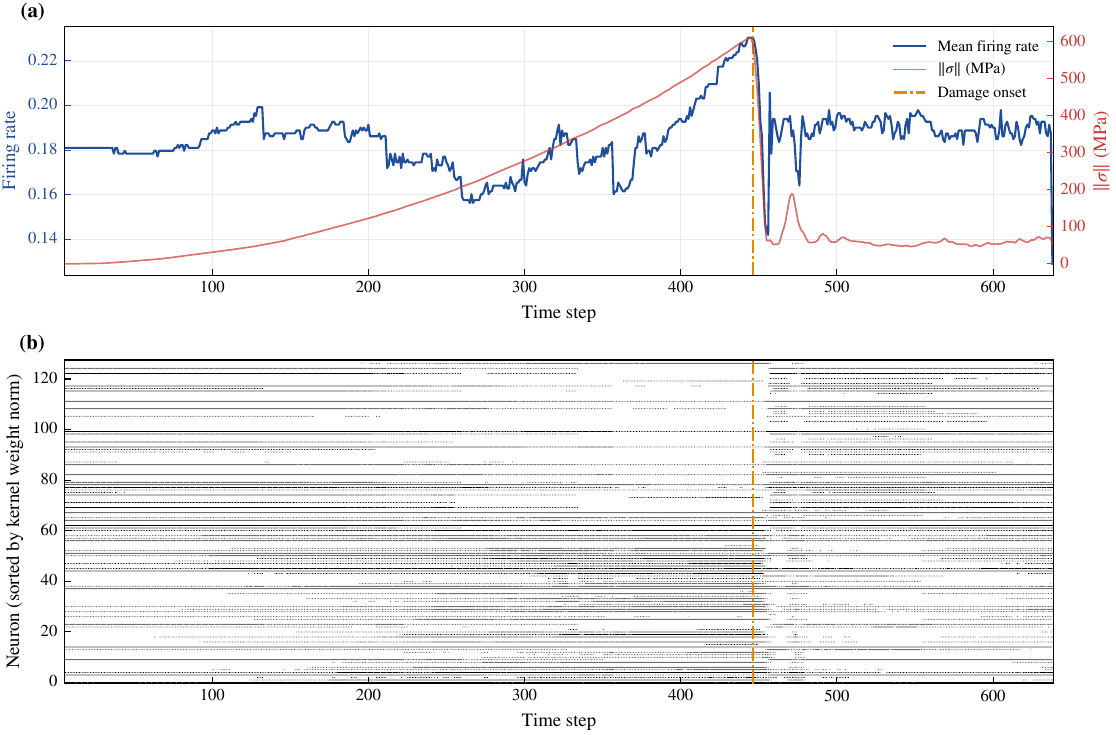}
\caption{SNN spiking patterns for an RVE under monotonic loading, hidden layer: A) mean SNN firing rate vs stress norm. B) SNN Neuron state vs time steps C) Neuron membrane potential evolution.}
\label{fig:Spike_L01}
\end{figure}
\begin{figure}[h]
\centering
\includegraphics[width=0.75\textwidth]{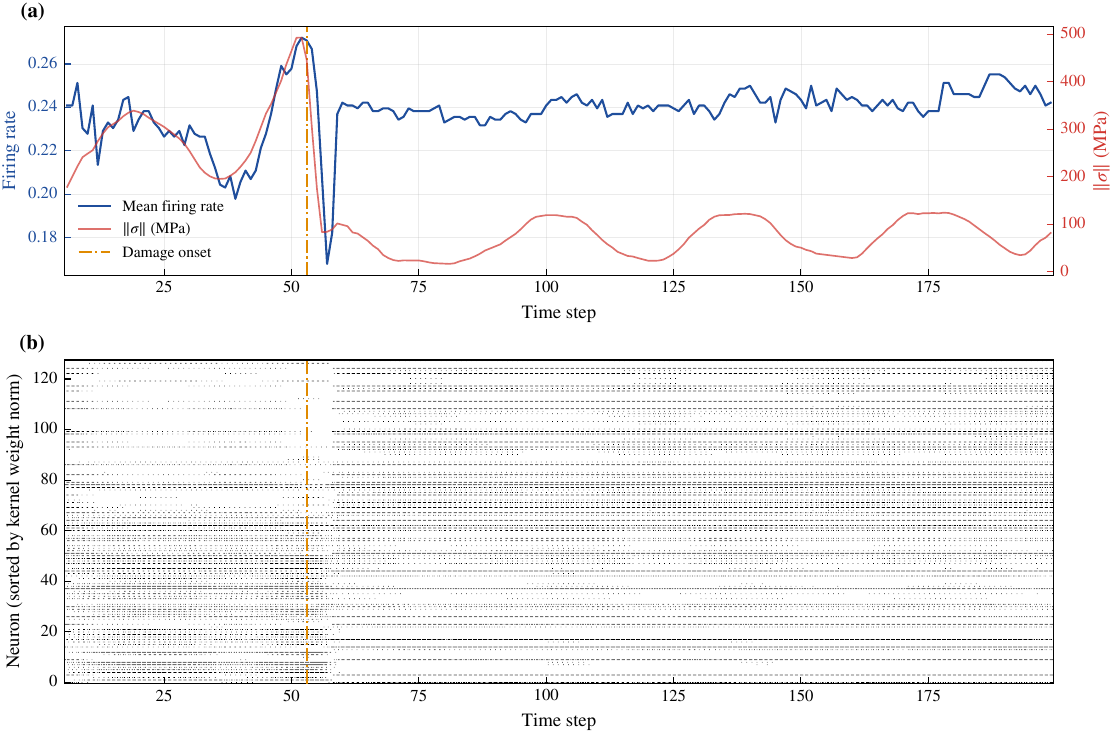}
\caption{SNN spiking patterns for an RVE under cyclic loading, hidden layer: A) mean SNN firing rate vs stress norm. B) SNN Neuron state vs time steps C) Neuron membrane potential evolution.}
\label{fig:Spike_L02}
\end{figure}
\end{document}